\documentclass[10pt,twocolumn,letterpaper]{article}

\usepackage[pagenumbers]{cvpr} 

\newcommand{\authorskip}{\qquad} 

\usepackage{graphicx}
\usepackage{amsmath}
\usepackage{amssymb}
\usepackage{booktabs}
\usepackage{textcomp}
\usepackage{gensymb}
\usepackage{booktabs}
\usepackage[table]{xcolor}
\usepackage{colortbl}
\usepackage{multirow}

\newtheorem{lemma}{LEMMA}
\usepackage[pagebackref=true,breaklinks=true,colorlinks,
citecolor=evaunit01green,urlcolor=evaunit01purple,bookmarks=false]{hyperref}
\usepackage{graphicx}
\usepackage{amsmath}
\usepackage{amssymb}
\usepackage{booktabs}

\usepackage[capitalize]{cleveref}
\crefname{section}{Sec.}{Secs.}
\Crefname{section}{Section}{Sections}
\Crefname{table}{Table}{Tables}
\crefname{table}{Tab.}{Tabs.}

\begin{document}

\title{Underwater object detection in sonar imagery with detection transformer and Zero-shot neural architecture search}

\author{XiaoTong Gu\textsuperscript{1}
\authorskip Shengyu Tang\textsuperscript{2}\thanks{Correspondence to:tangshengyu@ouc.edu.cn} 
\authorskip Yiming Cao\textsuperscript{1}
\authorskip Changdong Yu\textsuperscript{1} \\[0.5mm]
{
\fontsize{10.4pt}{9.84pt}\selectfont
\textsuperscript{1}Dalian Maritime University \hspace{5.5mm} \textsuperscript{2}Ocean Univesity of China
}\\
}
\hypersetup{
    colorlinks=true,
    linkcolor=red,    
    citecolor=red,    
    urlcolor=red      
}
\maketitle

\begin{abstract}
   Underwater object detection using sonar imagery has become a critical and rapidly evolving research domain within marine technology. However, sonar images are characterized by lower resolution and sparser features compared to optical images, which seriously degrades the performance of object detection.To address these challenges, we specifically propose a Detection Transformer (DETR) architecture optimized with a Neural Architecture Search (NAS) approach called NAS-DETR for object detection in sonar images. First, an improved Zero-shot Neural Architecture Search (NAS) method based on the maximum entropy principle is proposed to identify a real-time, high-representational-capacity CNN-Transformer backbone for sonar image detection. This method enables the efficient discovery of high-performance network architectures with low computational and time overhead. Subsequently, the backbone is combined with a Feature Pyramid Network (FPN) and a deformable attention-based Transformer decoder to construct a complete network architecture. This architecture integrates various advanced components and training schemes to enhance overall performance. Extensive experiments demonstrate that this architecture achieves state-of-the-art performance on two Representative datasets, while maintaining minimal overhead in real-time efficiency and computational complexity. Furthermore, correlation analysis between the key parameters and differential entropy-based fitness function is performed to enhance the interpretability of the proposed framework. To the best of our knowledge, this is the first work in the field of sonar object detection to integrate the DETR architecture with a NAS search mechanism.
\end{abstract}

\section{Introduction}\label{sec1}
Underwater object detection has widespread applications in ocean resource exploration, underwater military defense, underwater archaeology, and marine environmental monitoring~\cite{1,2,3,4}. Due to the complexity of the underwater environment, conventional optical imaging techniques perform poorly under high turbidity and low light conditions, making it difficult to obtain clear, high-quality underwater images. In contrast, sonar imaging systems leverage the low attenuation of sound waves in water, enabling effective surpassing of optical imaging limitations and facilitating real-time detection of underwater objects. However, the distinctive characteristics of sonar imaging lead to significant challenges, including severe noise interference, low resolution, small target sizes, and high inter-target similarity~\cite{5,6}. These factors considerably hinder the development of robust and accurate sonar image-based object detection algorithms.

At present, underwater object detection methods in sonar images can broadly be classified into two categories: knowledge-driven approaches and data-driven approaches. The knowledge-driven approach relies on prior knowledge of object attributes, such as geometric shapes, textural features, and spatial distribution patterns. It achieves object-background differentiation and localization through specialized morphological processing, clustering algorithms, and template matching strategies~\cite{7,8,9,10}. For example, Williams et al.~\cite{11} proposed a novel environmental feature-adaptive cascade algorithm for underwater sonar image detection. The algorithm leverages integral images to estimate background and shadow regions, identifying the target area based on pixel values from both. Finally, pseudo-targets are eliminated by incorporating the target’s physical structure, enabling accurate localization of the frogman. Cho et al.~\cite{12} proposed an underwater object detection algorithm using a sonar image simulator. The simulator generates template images at various angles, and the matching algorithm directly compares these templates with real sonar images, thereby overcoming the challenge of object recognition from a single angle. Ben et al.~\cite{13} proposed an underwater object segmentation algorithm based on the Markov Random Field model, achieving rapid target localization by distinguishing foreground objects from the background. Abu et al.~\cite{14} introduced a sonar image segmentation algorithm based on fuzzy kernel metrics, utilizing local spatial and statistical information as fuzzy terms to enhance background homogeneity while preserving texture features in regions of interest. This approach achieved optimal segmentation accuracy in multi-beam and synthetic aperture sonar images. Liu et al.~\cite{15} developed a two-stage sonar image small object detection approach that does not rely on prior sample information, effectively detecting small targets from low-quality side-scan sonar images. Yet, these knowledge-driven methods often rely on manual expertise and are computationally expensive, which significantly limits their performance and efficiency in practical applications.

In recent years, with the rapid development of deep learning technologies, object detection methods based on convolutional neural networks (CNNs) have achieved significant success across various domains~\cite{16,17,18,19}. In particular, the YOLO series of single-stage detection models~\cite{20,21,22,23} have gained widespread application due to their superior inference speed and accuracy. Hence, many researchers have adapted CNN-based object detection algorithms for sonar image detection. For example, Wang et al.~\cite{24} introduced a novel sonar image object detector called MLFFNet, which comprises multi-scale convolution modules, multi-level feature extraction modules, and multi-level feature fusion modules, achieving strong performance on public datasets. Shi et al.~\cite{25} proposed an advanced detection framework that integrates efficient feature extraction with multi-scale feature fusion modules, utilizing EfficientNet as the backbone network. By optimizing the network's depth, width, and resolution, EfficientNet demonstrates exceptional feature extraction capabilities. Additionally, several related research reviews have provided detailed descriptions and analyses of underwater object detection based on sonar images~\cite{5,6,26,27}. Although CNN-based methods for sonar object detection have achieved notable advancements in recent years, these approaches remain limited by constrained local receptive fields and insufficient capacity for nonlinear feature extraction. In contrast, the emerging DETR series of detectors have shown stronger competitiveness, can effectively model the contextual information of the objects, and enhance the ability of nonlinear feature extraction. However, the Transformer module in the DETR architecture directly adopts backbone designs originally intended for optical image classification tasks, with optimization objectives still confined to the characteristics of optical imagery. Compared to optical images, sonar images are inherently limited by low resolution and high levels of noise, posing substantial challenges to the performance of existing DETR-based frameworks on sonar data. Consequently, optimizing the DETR architecture in accordance with the unique characteristics of sonar imagery remains a critical and unresolved research issue.

Motivated by the above, we propose a novel sonar image object detection method called NAS-DETR. Specifically, we first apply the principle of maximum entropy and utilize an evolutionary algorithm to maximize the differential entropy of the neural network as the optimization objective. This approach aims to search for a more efficient CNN-Transformer hybrid backbone, which serves as the foundation for constructing the overall network architecture. Based on the hybrid backbone architecture, NAS-SDETR is developed by integrating a Feature Pyramid Network (FPN) with a deformable-attention based transformer decoder. By incorporating a novel content-position decoupled query initialization strategy and a multi-task optimized hybrid loss function, NAS-DETR significantly improves the robustness in low-resolution and high-noise sonar image scenarios. The decoder incorporates an adaptive mechanism that fuses multi-scale features and dynamically updates anchor box offsets, effectively addressing the limitations of traditional DETR models in terms of slow convergence and insufficient localization accuracy. Furthermore, we are the first to propose a differential entropy–ar-chitecture parameter correlation analysis method based on architecture parameter averaging. By leveraging the Spearman rank correlation coefficient, we quantitatively reveal the monotonic dependencies between differential entropy and key architectural parameters—such as the depth and width of the CNN-Transformer backbone and the hidden dimension of the Transformer. The proposed method introduces theoretical interpretability into neural architecture search and provides statistical support for further optimization in sonar object detection tasks. In summary, the main contributions of this work are as follows:
\begin{itemize}
   \item Based on the maximum entropy principle, we maximize the differential entropy of the neural network as the optimization goal and search for an efficient CNN-Transformer hybrid network backbone.
   \item By introducing a content-position decoupled query initialization strategy and a multi-task hybrid loss function into the NAS-DETR architecture, the model’s robustness to sonar images in low-resolution and high-noise scenarios is significantly enhanced.
   \item We proposed a differential entropy-architecture parameter correlation analysis method based on architecture parameter averaging, which introduced interpretability theory support for NAS. 
   \item Extensive experiments are performed to demonstrate that our proposed method achieves state-of-the-art performance on different datasets.
\end{itemize}

The rest of this paper is structured as follows: Section~\ref{sec2} reviews related work on DETRs and NAS methods. Section~\ref{sec3} details the principles and architecture of the proposed approach. Section~\ref{sec4} presents the evaluation results, followed by a brief conclusion in Section~\ref{sec5}.

\section{Related work}\label{sec2}
Since our work primarily draws on object detection methods based on Detection Transformers (DETRs) and Neural Architecture Search (NAS) strategies, this section mainly describes the related work on DETRs and NAS.
\subsection{DETRs}
YOLO-based detectors have gained popularity among researchers due to their excellent trade-off between accuracy and real-time performance~\cite{18,19,20,21}. However, these CNN-based detectors typically require post-processing using non-maximum suppression (NMS), which can result in instability in both speed and accuracy.

In recent years, DETRs have made significant progress in the field of object detection due to their streamlined architectural design and reduced reliance on handcrafted components~\cite{28,29,30,31}. For instance, Carion et al.~\cite{28} first proposed a Transformer-based encoder-decoder architecture, framing object detection as a direct set prediction problem. This approach simplifies the detection pipeline and effectively eliminates the need for many manually designed components. Liu et al.~\cite{29} proposed a novel query mechanism for DETR, utilizing dynamic anchor boxes that iteratively update coordinates layer-by-layer within the Transformer decoder, aiming to improve detection performance and refine query representation. Zhu et al.~\cite{30} integrated a deformable attention module into DETR, enabling the model to concentrate on a sparse set of key sampling points around reference points, thereby addressing DETR’s slow convergence and limited feature resolution issues. Huang et al.~\cite{31} proposed a novel training scheme that leverages bounding boxes predicted by multiple teacher detectors to achieve efficient knowledge distillation for the DETR object detector. Despite the significant advantages of DETR, it is constrained by several limitations, including slow training convergence, substantial computational resource demands, and difficulties in optimizing the query mechanism. To mitigate these issues, a variety of DETR variants have been developed~\cite{32,33,34,35,36}.
For instance, Li et al. conducted a comprehensive analysis of the factors contributing to the slow convergence of DETR-like methods and introduced an innovative denoising training strategy to expedite the training process of DETR models~\cite{32}. Yao et al.~\cite{33} and Roh et al.~\cite{34} successively reduced computational costs by decreasing the number of encoder and decoder layers or updating the number of queries. Wang et al.~\cite{35} proposed a novel anchor-based query design for Transformer-based object detection to address the issues of the traditional DETR query mechanism, such as the lack of clear physical meaning and optimization challenges. While DETR variants have increasingly gained attention from researchers due to their advantages, to the best of our knowledge, no Transformer-based detector has been specifically designed for underwater object detection in sonar images.

\subsection{NAS}
Neural Architecture Search (NAS) automates the identification of optimal network architectures, significantly reducing the reliance on manual design efforts and enhancing both the efficiency and performance of model development~\cite{36}. As a result, NAS plays a vital role in object detection tasks and remains an active area of research. For instance, Wang et al.~\cite{37} developed an efficient NAS framework specifically tailored for object detection tasks, which focuses on optimizing the decoder architecture to substantially reduce both manual design efforts and computational overhead. Yao et al. proposed a two-stage NAS framework, termed SM-NAS, to optimize the overall architecture design of object detection models, achieving an improved trade-off between computational cost and detection accuracy~\cite{38}. Li et al.~\cite{39} introduced a novel FPN search framework, named AutoDet, which autonomously identifies information connections across multi-scale features and configures detection architectures with high efficiency and state-of-the-art performance. Although one-shot NAS methods have demonstrated promising performance, the reliance on weight sharing within a supernet for architecture evaluation leads to substantial evaluation bias and high training overhead. In contrast, Zero-shot NAS achieves rapid performance prediction without training by analyzing prior characteristics intrinsic to the architecture, thereby offering superior efficiency and scalability~\cite{40,41,42}. For example, Sun et al. proposed a Zero-shot NAS approach named MAE-DET for object detection, which leverages the principle of maximum entropy to automatically design efficient detection backbones without requiring network parameter training, thereby reducing the architecture design cost to nearly zero~\cite{41}. Shen et al. introduced a purely mathematical framework, DeepMAD~\cite{42}, for discovering high-performance convolutional neural networks. The core idea of DeepMAD is to maximize network entropy while maintaining the validity of the network at a level close to a small constant. Inspired by these studies, we adopt a Zero-shot NAS method grounded in the maximum entropy principle to efficiently identify a CNN-Transformer backbone with strong representational capacity for sonar image detection.

\section{Porposed NAS-DETR framework}\label{sec3}
\begin{figure*}
	\centering
	\includegraphics[width=0.95\linewidth]{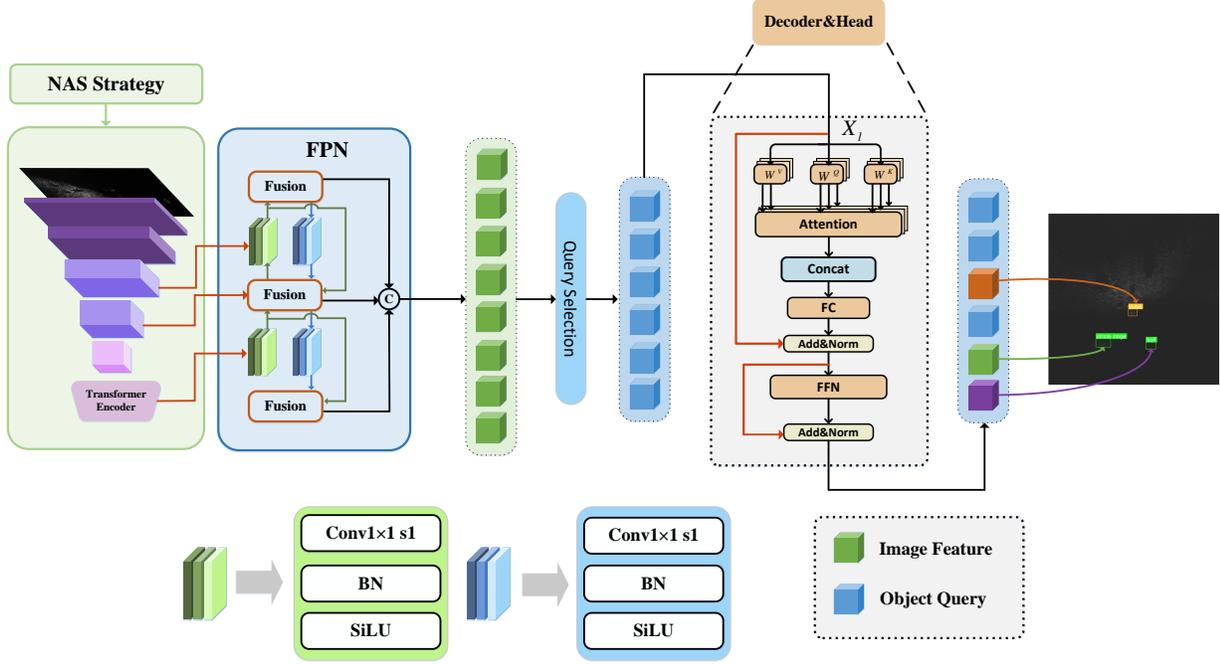}
	\caption{The proposed NAS-DETR framework for sonar image object detection.}\label{fig1}
\end{figure*}

Figure~\ref{fig1} displays an overview of the NAS-SDETR architecture, which contains the neural architecture search mechanism and the improved DETR structure. For NAS, we leverage a Zero-shot architecture search method grounded in the maximum entropy principle to identify the optimal backbone network. In this way, this approach enables the DETR architecture to accommodate a wider spectrum of data distributions, such as sonar images, rather than being limited to optical imagery. An evolutionary algorithm is employed as the optimization strategy, where individual network architectures undergo continuous mutation. Individuals with higher architectural differential entropy are retained, while those with low fitness scores or exceeding a predefined FLOPs threshold are eliminated to update the population iteratively. To enhance the architectural expressiveness and search efficiency, we introduce an innovative feature extraction backbone that seamlessly integrates CNN and Transformer encoders. This unified design enables Transformer blocks to be directly involved in the NAS process, facilitating more effective architecture exploration. Downstream components include a Feature Pyramid Network (FPN), a query selection module, and a Transformer decoder equipped with deformable attention. The multi-scale features produced by the CNN-Transformer backbone are subsequently fused and decoded through an encoder-decoder architecture, ultimately yielding high-precision detection outcomes.

\subsection{Entropy of CNN-Transformer backbone} \label{sec:3.1}
According to~\cite{44}, the upper bound of a neural network’s differential entropy imposes a constraint on the theoretical maximum information capacity of its feature space, thereby influencing the network’s representation capability. In other words, the differential entropy is positively correlated with the network’s representational performance, such as classification accuracy. To this end, we search for CNN-Transformer backbone architectures that maximize output differential entropy, aiming to identify the optimal design. To enable quantitative analysis, the neural network is subsequently formulated as an information processing system.

First, we model a neural network as a graph structure~\cite{44} as follows:
\begin{equation}
G=(v, e)    
\end{equation}
For any $v \in \mathcal{V}$ and $e \in \mathcal{E}$, $h(v) \in \mathbb{R}$ and $h(e) \in \mathbb{R}$ present the Differential Entropy values endowed with each vertex $v$ and each edge $e$. Therefore, the continuous state space $S$ of the neural network architecture is represented as:
\begin{equation}
S=\{h(v), h(e):\forall v \in \mathcal{V}, e \in \mathcal{E}\}
\end{equation}
The reasoning process of the neural network is expressed as:
\begin{equation}
F^{\prime}=G(F)    
\end{equation}
where $F$ represents the input of the entire architecture, and $F^{\prime}$ is the output feature map.

In the graph structure of the neural network, the vertex set $S_{v}=\{v: \forall v \in \mathcal{V}\}$ consists of neurons, and the edge set $S_{e}=\{e: \forall e\in \mathcal{E}\}$ consists of signal channels between neurons. We denote the components of the total differential entropy on these two parts as $H\left(S_{v}\right)=\{h(v): \forall v \in \mathcal{V}\}$ and $H\left(S_{e}\right)=\{h(e): \forall e \in \mathcal{E}\}$. In the Zero-shot NAS process, we only intervene in the $S_v$ set (i.e., only change the structural arrangement), and use the gradient descent algorithm to optimize $S_e$ after determining $S_v$. Hence, in the rest of this work, the differential entropy of the neural network is assumed to refer to $H\left(S_{v}\right)$. However, for deep neural networks, directly computing $H\left(S_{v}\right)$ is intractable. When both $F$ and $S_e$ are assumed to follow a standard Gaussian distribution $N(0,1)$, the following relation holds:
\begin{equation}
H\left(S_{v}\right) \approx H\left(F^{\prime}\right)    
\end{equation}
Therefore, the easier-to-solve $H\left(F^{\prime}\right)$ can be used to approximate $H\left(S_{v}\right)$ for calculation. However, it is still unrealistic to directly give the analytical formula for the differential entropy of the feature graph $F^{\prime}$. We turn to optimize the network structure by maximizing the upper bound of $H\left(F^{\prime}\right)$. The lemma 1 is introduced as follows:
\begin{lemma}
(Maximum Differential Entropy).
Let $x$ be a continuous random variable with probability density function $p(x)$, mean $\mu$, and variance $\sigma^{2}$. The differential entropy of $X$, defined as:
\begin{equation}
h(x)=-\int_{-\infty}^{\infty} p(x) \ln p(x) d x    
\end{equation}
satisfies the inequality:
\begin{equation}
h(x) \leq \frac{1}{2} \ln \left(2 \pi e \sigma^{2}\right)    
\end{equation}
with equality if and only if $x$ follows a Gaussian distribution, i.e.,
\begin{equation}
p(x)=\frac{1}{\sqrt{2 \pi \sigma^{2}}} e^{-\frac{(x-\mu)^{2}}{2 \sigma^{2}}} \end{equation}
\end{lemma}
The $\mu$ and $\sigma^{2}$ of the output feature map $F^{\prime}$ are easy to obtain by statistical methods. It is not difficult to find that the upper bound is absolutely positively correlated with the standard deviation $\sigma$. The following is an analytical representation of the standard deviation of the feature map of the CNN-Transformer hybrid architecture.

For Zero-shot NAS mechanisms based on the principle of maximum entropy, nonlinear operations such as ReLU, Softmax, and LayerNorm can significantly affect the estimation of the standard deviation of feature maps. In contrast, convolution operations can be considered equivalent to matrix multiplication in optimized computation, and thus CNN layers can be regarded as linear operations~\cite{43}. Taking the output of a single channel in the feature map as an example, let the input feature map be denoted as $X$, and the output feature map as $Y$. All learnable parameters can be represented as a matrix $M$. Consider a row vector $M_{i} \in M$, where $M_{i} \in \mathbb{R}^{C \times K^{2}}$. Let $m_{i j} \in M_{i}$ denote an arbitrary element in this vector. Similarly, extract a column vector $x_{k} \in X$, where $x_{k} \in \mathbb{R}^{C \times K^{2}}$, and let $x_{k} \in \mathbb{R}^{C \times K^{2}}$ be an arbitrary element in this vector. Then, the following relationship holds:
\begin{equation}
\left\{\begin{array}{c}
m_{i j} \sim N(0,1) \\
x_{j k} \sim N(0,1)
\end{array}\right.    
\end{equation}
For any element $y_{i k} \in Y$, then there is:
\begin{equation}
y_{i k}=\sum_{j=1}^{K^{2} \times C} m_{i j} x_{j k}
\end{equation}

To facilitate the subsequent derivation, we introduce Lemma 2, which is described as follows:
\begin{lemma}
Suppose matrices $A$ and $B$ contain elements $a_{i j}$ and $b_{j k}$, respectively, where each element is independently drawn from a distribution with 
$\mu=0$ and $\sigma=1$. Assume that the number of columns in $A$ and the number of rows in $B$ are both equal to $S$. Then, $\sigma^{2}\left(\sum_{j=1}^{S} a_{i j} b_{j k}\right)=\sum_{j=1}^{S} \sigma^{2}\left(a_{i j}\right) \sigma^{2}\left(b_{j k}\right)$.
\end{lemma}
It can be observed that if the output feature map has a size of $(w, h)$, then $i \in(0, w)$ and $j \in(0, h)$. By further incorporating the rotational symmetry of individual elements and Lemma 2, the statistical properties of the output feature elements can be derived as follows:
\begin{equation}
\begin{aligned}
\mathbb{E}\left(y_{i k}\right) & =\mathbb{E}\left(\sum_{j=1}^{K^{2} \times C} M_{i j} x_{j k}\right) \\
& =\sum_{j=1}^{K^{2} \times C} \mathbb{E}\left(M_{i j} x_{j k}\right) \\
& =\sum_{j=1}^{K^{2} \times C} \mathbb{E}\left(M_{i j}\right) \mathbb{E}\left(x_{j k}\right)\\
&=0
\end{aligned}    
\end{equation}
\begin{equation}
\begin{aligned}
\sigma^{2}\left(y_{i k}\right) & =\sigma^{2}\left(\sum_{j=1}^{K^{2} \times C} M_{i j} x_{j k}\right) \\
& =\sum_{j=1}^{K^{2} \times C}\left\{\sigma^{2}\left(M_{i j}\right) \sigma^{2}\left(x_{j k}\right)\right\} \\
& =\sum_{j=1}^{K^{2} \times C} \sigma^{2}\left(x_{j k}\right) \\
& =K^{2} C \sigma^{2}\left(x_{j k}\right)
\end{aligned}
\end{equation}
For a convolutional neural network (CNN) with $L$ layers, the value of any output feature map element can be expressed as:
\begin{equation}
\mathbb{E}\left(y_{i k}^{L}\right)=0    
\end{equation}
\begin{equation}
\sigma^{2}\left(y_{i k}^{L}\right)=\prod_{l=1}^{L} K_{L}^{2} C_{L}
\end{equation}
Based on the above recursive relationship, it can be concluded that both the input feature maps and model parameters follow a standard Gaussian distribution. The differential entropy of a single output element is given by:
\begin{equation}\label{eq:14}
H_{L}=\sum_{i=1}^{L} \log \left(c_{i} k_{i}^{2}\right)
\end{equation}

Next, we present the recursive formulation of variance propagation within the Transformer block. The operations of the Transformer block can be described by the following equations:
\begin{equation}
\left\{\begin{array}{l}
Q=X W^{Q} \\
K=X W^{K} \\
V=X W^{V}
\end{array}\right.    
\end{equation}
\begin{equation}
\left\{\begin{array}{l}
A=\operatorname{Softmax}\left(\frac{Q K^{T}}{\sqrt{d_{k}}}\right) \\
H=A V \\
Y=H M_{1} M_{2}
\end{array}\right.
\end{equation}
where $X$ is the input feature map, $W^{Q, K, V}$ is the attention input projection parameter matrix, $A$ is the attention map, $H$ is the Attention output projection parameter matrix, $M_1$, $M_2$ are the FFN layer parameter matrices. Here, we only consider the final two-dimensional shape of each matrix, and Table~\ref{tab:t1} gives the parameters of each variable.
\begin{table}[htb] 
	\centering
	\caption{Transformer Block tensors and sizes. \label{tab:t1}}
	\setlength{\tabcolsep}{3mm}
        \renewcommand\arraystretch{1.15}{
       \begin{tabular}{@{}cc@{}}
		\toprule
Tensor             & Size \\  \midrule
$X$                & $\left[\mathrm{S}, d_{\text {model}}\right]$                \\
$W^{Q, K, V}$      & $\left[d_{\text {model }}, d_{\text {model }}\right]$       \\
$A$                & $[S, S]$                                                     \\    
$H$                & $\left[\mathrm{S}, d_{\text {model }}\right]$         \\ 
$M_{1}$            &  $\left[d_{\text {model }}, d_{\text {FFN }}\right]$  \\
$M_{2}$            &  $\left[d_{\text {FFN }}, d_{\text {model }}\right]$ \\\bottomrule                                   
	\end{tabular}}
\end{table}
The following derivation is similar to that of CNN, and the details are not repeated here. The projection process of $V$ is described as follows:
\begin{equation}
\begin{aligned}
\sigma^{2}\left(V_{j k}\right) & =\sigma^{2}\left(\sum_{l=1}^{d_{\text {model }}} x_{j l} W_{l k}^{V}\right) \\
& =d_{\text {model }} \sigma^{2}\left(x_{j l}\right)
\end{aligned}    
\end{equation}
It is not difficult to observe that when $a_{i j} \in A$, we have $a_{i j} \in(0,1)$, and thus $\sigma^{2}\left(a_{i j}\right)<1$. This condition is employed in the following derivation for scaling purposes. In essence, this treatment is equivalent to neglecting the non-linear Softmax operation. According to Lemma 2, under purely linear operations, the expectation is transitive across layers and remains zero, i.e., $\mathbb{E}\left(a_{i j}\right)=\mathbb{E}\left(V_{j k}\right)=0$. Based on this condition, it is straightforward to derive the following:
\begin{equation}
\sigma^{2}\left(H_{i j}\right)=\sigma^{2}\left(\sum_{l=1}^{S} a_{i j} V_{j k}\right)    
\end{equation}
\begin{equation}
\sigma^{2}\left(\sum_{l=1}^{S} a_{i j} V_{j k}\right)<S \sigma^{2}\left(V_{j k}\right)
\end{equation}
\begin{equation}
S \sigma^{2}\left(V_{j k}\right)=d_{\text {model }} S \sigma^{2}\left(x_{j l}\right)
\end{equation}
Finally, $H$ passes through two linear layers to get the final output. Since the linear layer and the convolutional layer mentioned above have the essence of matrix multiplication, we can infer the following formula:
\begin{equation}
\begin{aligned}
\sigma^{2}\left(y_{i j}\right) & =F F N_{\times 2}\left(\sigma^{2}\left(H_{i j}\right)\right) \\
& =d_{\text {model }} d_{\text {feedforward }} \sigma^{2}\left(H_{i j}\right) \\
& =d_{\text {model }}^{2} d_{\text {feedforward }} S^{2}\left(x_{j l}\right)
\end{aligned}    
\end{equation}
Based on the above analysis, the differential entropy of a single element in the output feature map of the final layer of the CNN-Transformer hybrid backbone can be expressed as follows:
\begin{equation}\label{eq:22}
H_{L}=d_{\text {model }}^{2} d_{\text {feedforward }} S \sum_{i=1}^{L} \log \left(c_{i} k_{i}^{2}\right)    
\end{equation}
The above process fully describes the search object of the NAS-DETR method, the standard deviation of the CNN-Transformer Backbone and the representation method of differential entropy. Next, we will propose a NAS search strategy based on this method.

\begin{figure}
	\centering
	\includegraphics[width=0.95\linewidth]{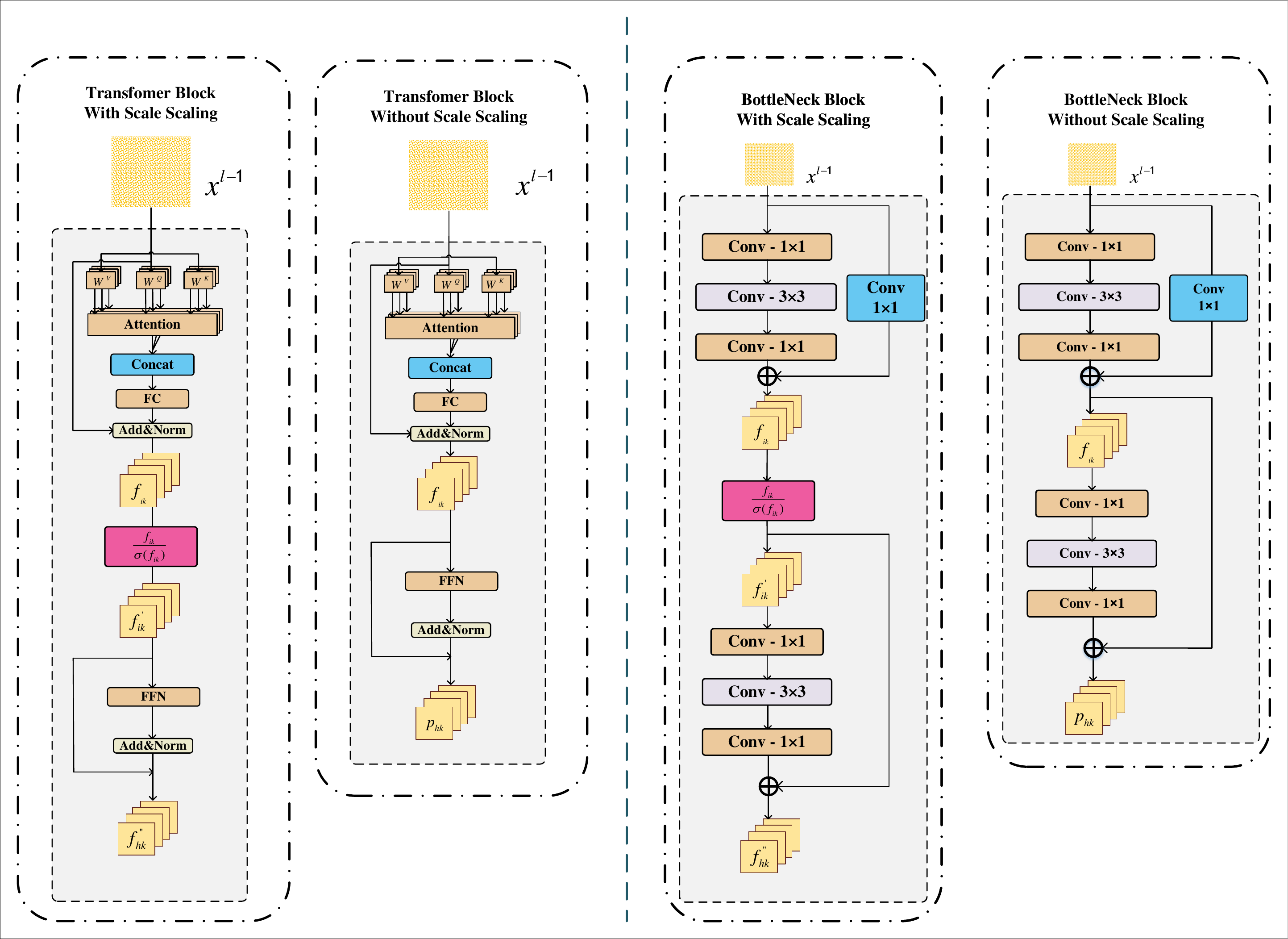}
	\caption{Schematic diagram of the two feedforward modes under the Transformer Block and BottleNeck module.}\label{fig2}
\end{figure}

\begin{figure*}
	\centering
	\includegraphics[width=0.88\linewidth]{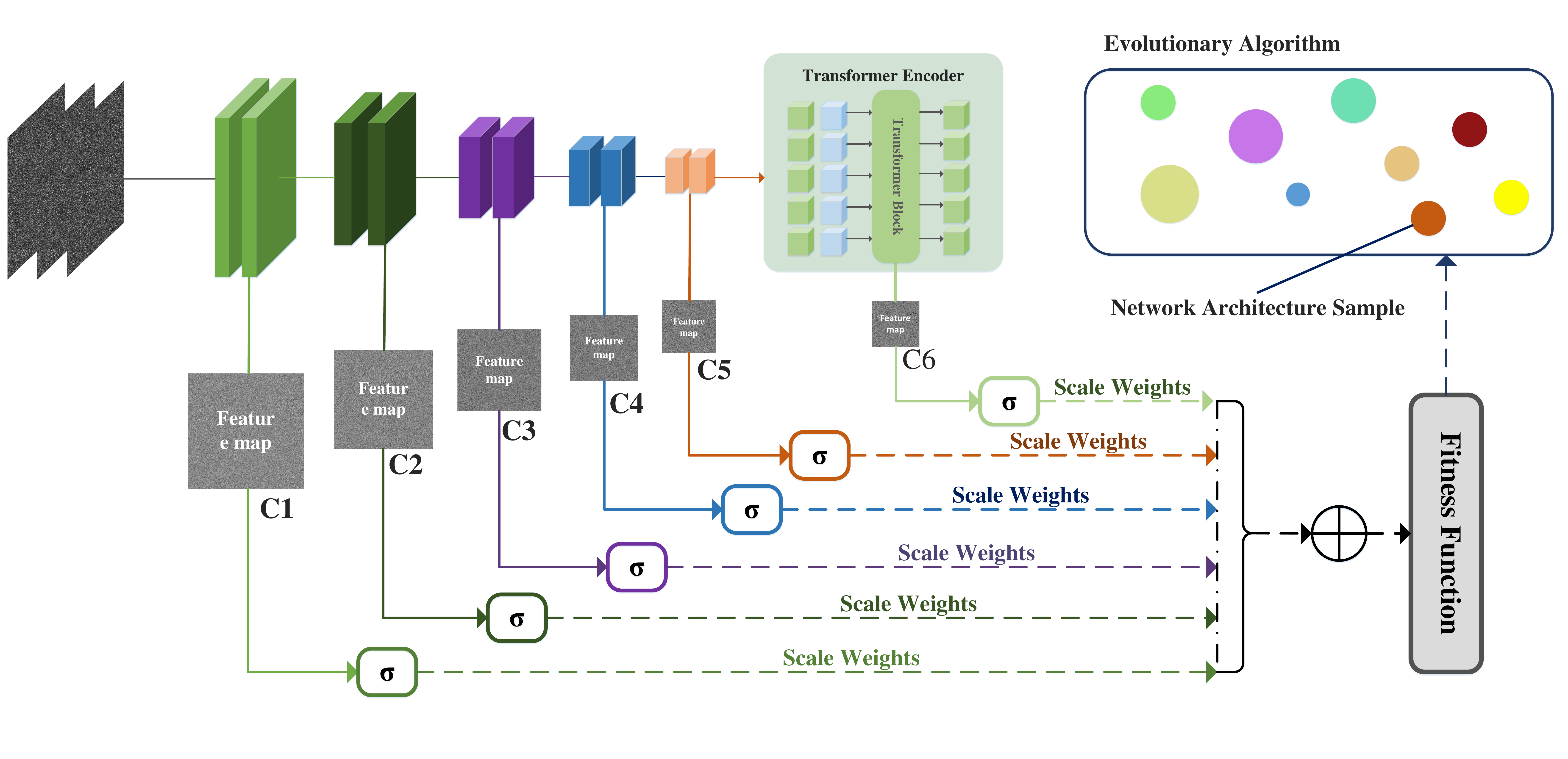}
	\caption{Schematic diagram of the proposed architecture search strategy and fitness function calculation.}\label{fig3}
\end{figure*}

\subsection{NAS Strategy} 
In this section, we present a specific engineering-based estimation scheme for differential entropy during the search process (see Figure~\ref{fig3}), the construction of the fitness function, and the use of evolutionary algorithms (EAs) as solvers to perform architecture search. To eliminate the influence of uncertainties such as the framework and computational hardware, we inject Gaussian noise repeatedly into the neural network inputs and compute the average of the variance statistics of multiple output feature maps to estimate the differential entropy.

While this method improves the estimation accuracy, it also introduces a critical issue—namely, in deep neural networks, feature values may be amplified or attenuated across layers due to the stochastic nature of weight initialization. This can result in numerical overflow (gradient explosion) or underflow (gradient vanishing), thereby undermining the Gaussian assumption. To address this, we propose a numerical scale normalization method, with the detailed procedure described below.

We scale the intermediate feature maps of Transformer Block and BottleNeck respectively in Figure~\ref{fig2}. The execution position is similar to the position of BN in the original structure. The scaling operation is expressed as:
\begin{equation}\label{eq:23}
F^{\prime}=\frac{F}{\sigma(F)}    
\end{equation}

Let $F$ and $F^{\prime}$ denote the feature maps before and after scaling, respectively. Although this scaling operation effectively mitigates the aforementioned numerical instability, it introduces a new issue: the standard deviation of the current feature map, appearing in the denominator of Eq.\eqref{eq:23}, affects the variance propagation discussed in Section \ref{sec:3.1}. Consequently, when $F$ is processed through two consecutive convolutional layers, the estimated variance of the final output feature map deviates from the actual value.

To quantify this discrepancy, we define $P$ as the output of two convolutional layers without scaling, and $F^{\prime \prime}$ as the output with scaling applied. Let $f_{i k} \in F$, $f_{i k}^{\prime} \in F^{\prime}$, $f_{h k}^{\prime \prime} \in F^{\prime \prime}$, and $p_{h k} \in P$ be individual elements of their respective feature maps. Let $M^{A}$ and $M^{B}$ denote the weight matrices of the two convolutional layers. The kernel sizes and channel numbers of the two layers are represented by $K_{1}$, $K_{2}$ and $C_{1}$, $C_{2}$, respectively. The variance propagation issue described above is quantitatively characterized as follows:
\begin{equation}
\left\{\begin{array}{l}
f_{i k}=\sum_{j=1}^{K_{1}^{2} \times C_{1}} M_{i j}^{A} x_{j k} \\
f_{i k}^{\prime}=\frac{f_{i k}}{\sigma\left(f_{i k}\right)} \\
f_{h k}^{\prime \prime}=\sum_{i=1}^{K_{2}^{2} \times C_{2}} M_{h i}^{B} f_{i k}^{\prime} \\
p_{h k}=\sum_{i=1}^{K_{2}^{2} \times C_{2}} M_{h i}^{B} f_{i k}
\end{array}\right.
\end{equation}
The above equations correspond to the two reasoning processes in Figure~\ref{fig2}, and it is easy to get:
\begin{equation}
\boldsymbol{\sigma}^{2}\left(p_{h k}\right) \neq \boldsymbol{\sigma}^{2}\left(f_{h k}^{\prime \prime}\right)    
\end{equation}
Therefore, we propose an effective variance estimation scheme to solve this problem as follows.
\begin{equation}
{\sigma}^{2}\left(p_{h k}\right)={\sigma}^{2}\left(f_{i k}\right) {\sigma}^{2}\left(f_{h k}^{\prime \prime}\right)
\end{equation}
The effectiveness of this scheme is demonstrated as follows:

According to LEMMA2, $M_{i j}^{A}$ and $M_{h i}^{B}$ follow standard Gaussian distributions. Hence, 
\begin{equation}
\begin{aligned}
{\sigma}^{\mathbf{2}}\left(p_{h k}\right) & ={\sigma}^{\mathbf{2}}\left(\sum_{i=1}^{K_{2}^{2} \times C_{2}} M_{h i}^{B} f_{i k}\right) \\
& =\sum_{i=1}^{K_{2}^{2} \times C_{2}} {\sigma}^{\mathbf{2}}\left(M_{h i}^{B}\right) {\sigma}^{\mathbf{2}}\left(f_{i k}\right) \\
& =\sum_{i=1}^{K_{2}^{2} \times C_{2}}\left[{\sigma}^{\mathbf{2}}\left(M_{h i}^{B}\right) {\sigma}^{\mathbf{2}}\left(\frac{f_{i k}}{\sigma\left(f_{i k}\right)}\right)\right] {\sigma}^{\mathbf{2}}\left(f_{i k}\right) \\
& =\left[{\sigma}^{\mathbf{2}}\left(\sum_{i=1}^{K_{2}^{2} \times C_{2}} M_{h i}^{B} f_{i k}^{\prime}\right)\right] {\sigma}^{\mathbf{2}}\left(f_{i k}\right) \\
& ={\sigma}^{\mathbf{2}}\left(f_{i k}\right) {\sigma}^{\mathbf{2}}\left(f_{h k}^{\prime \prime}\right)
\end{aligned}
\end{equation}

On the other hand, during the practical architecture search process, we observed that if differential entropy alone is used as the fitness score in the evolutionary algorithm, the number of layers $L$ tends to dominate the score entirely. This phenomenon is further discussed in Section \ref{sec:4.6}. However, based on prior knowledge in neural network architecture design ~\cite{43}, it is well known that excessively deep networks may hinder effective information propagation, potentially leading to issues such as vanishing gradients.

Therefore, when designing the fitness function, it is essential to incorporate this prior by encouraging architectures with appropriate width. To this end, we define a modified fitness score $Z^{\prime}$, which jointly considers both the differential entropy (reflected in the feature map variance) and the allocation of reasonable layer width. Here, $h^{D}$ denotes the input feature map, and $C_{\mathrm{in}}$ represents the input channel number or vector dimension of a given block:
\begin{equation}
Z^{\prime}\left(h^{D}\right)=\log \left(\operatorname{var}\left(h^{D}\right)\right)+\log \left(C_{i n}\right)    
\end{equation}

In recent works from the DETR series ~\cite{45}~\cite{46}, multi-scale features are commonly utilized to improve the recall rate of the detector across objects of varying sizes. A typical approach involves using feature maps from the $C_{3}$, $C_{4}$, and $C_{5}$ stages of the backbone as inputs to the subsequent decoder.

Moreover, the amount of information (i.e., entropy) contained within intermediate feature maps significantly affects the network's feature representation capability. This insight motivates us to incorporate the average differential entropy across multiple scales into the fitness function $Z$. The computation process is defined in Eq. \eqref{eq:29}, where $a_{1}, a_{2}, \ldots, a_{6}$ serve as weighting coefficients to balance the contributions of differential entropy at different scales. A visual illustration of this process is provided in Figure~\ref{fig3}:
\begin{equation}\label{eq:29}
\begin{aligned}
{Z}({G})= & \frac{\mathbf{1}}{\mathbf{6}}\left\{a_{1} Z^{\prime}\left(C_{1}\right)+a_{2} Z^{\prime}\left(C_{2}\right)+a_{3} Z^{\prime}\left(C_{3}\right)\right. \\
& \left.+a_{4} Z^{\prime}\left(C_{4}\right)+a_{5} Z^{\prime}\left(C_{5}\right)+a_{6} Z^{\prime}\left(C_{6}\right)\right\}
\end{aligned}   
\end{equation}

\begin{figure}
	\centering
	\includegraphics[width=0.95\linewidth]{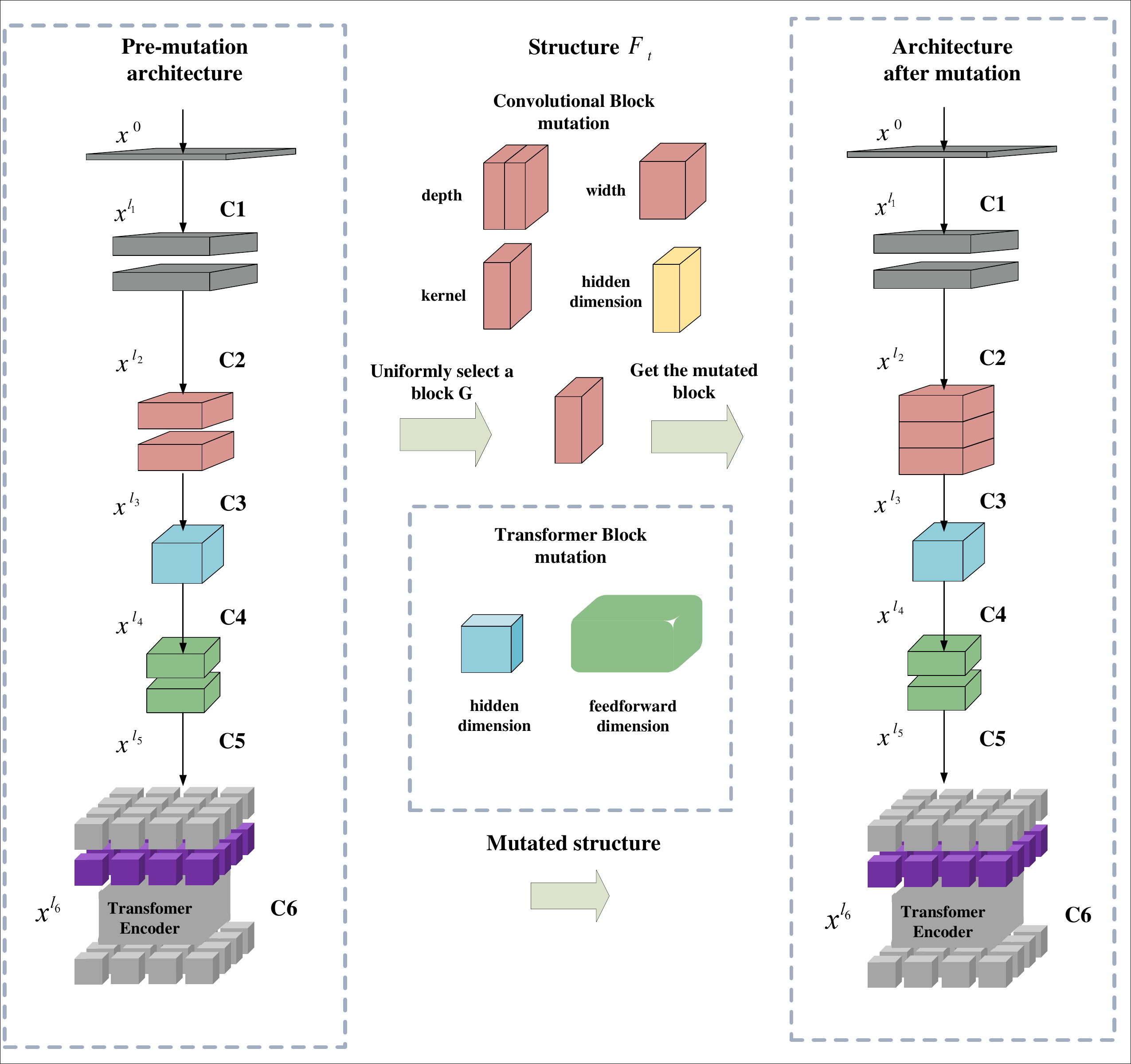}
	\caption{Mutation operation of the CNN-Transformer hybrid backbone network module.}\label{fig4}
\end{figure}

Finally, we adopt an evolutionary algorithm as the optimization strategy to maximize the fitness function. The algorithm begins by initializing a population of $N$ network architectures, with an upper bound imposed on FLOPs. Through iterative evolution, new individuals are generated by randomly modifying portions of the network architecture. The fitness of each new individual is then evaluated, and the population is updated by selecting the top-$N$ individuals with the highest fitness scores, while discarding those exceeding the FLOPs constraint.

We define each random modification of the network's architectural parameters as a mutation operation. In each iteration, four out of the six blocks in the network are randomly selected, and each of these four blocks undergoes two independent mutation operations. The mutation process in a single iteration is illustrated in Figure~\ref{fig4}. Detailed experimental settings and hyperparameters will be discussed in Section \ref{sec4}.

\begin{figure*}
	\centering
	\includegraphics[width=0.85\linewidth]{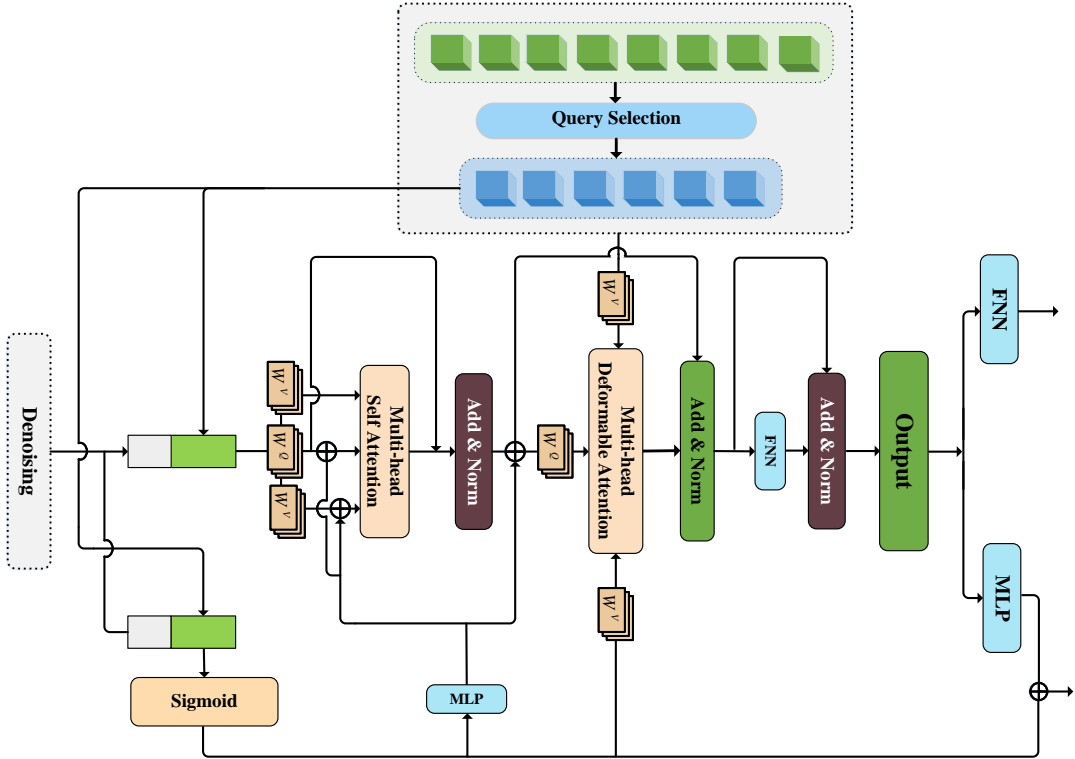}
	\caption{Architectural design of the Decoder and Head components.}\label{fig5}
\end{figure*}

\subsection{Decoder for Detectors} 
As illustrated in Figure~\ref{fig5}, the decoder of NAS-DETR incorporates several well-established design components that have been widely validated in prior works ~\cite{47}. The overall architecture closely follows that of the Deformable-DETR decoder , where the attention mechanism adopts multi-scale deformable attention to reduce computational complexity while effectively leveraging multi-scale features. The DETR family formulates object detection as a set prediction problem, aiming to find the optimal bipartite matching between a set of queries and the set of ground-truth objects. In our work, we follow Deformable-DETR's modeling of query semantics—dividing each query into a content query that encodes object-related information and a positional query that serves as an anchor. Given this formulation, the quality of the queries has a significant impact on the final detection accuracy. To ensure high-quality initial queries for the decoder, we incorporate a query selection mechanism ~\cite{48}, which selects features with high classification scores. A large number of features produced by the CNN-Transformer backbone are thus filtered through query selection to obtain a small set of high-quality queries, which are then used as inputs to the decoder.

During inference, if the matching between an individual anchor in the positional query and a ground truth object has been established, the decoding process of each decoder layer can be interpreted as predicting the relative offset between the ground truth object and the initial anchor. This is formally expressed in Eq. \eqref{eq:30}:
\begin{equation}\label{eq:30}
\left(x^{\prime}, y^{\prime}, w^{\prime}, h^{\prime}\right)=(x, y, w, h)+(\Delta x, \Delta y, \Delta w, \Delta h)    
\end{equation}

The above equation provides a physical interpretation of the decoder process. Taking the parameter $x$ as an example: $x$ denotes the initial position parameter of the anchor box provided by the query. It serves as one of the inputs to a decoder layer, which predicts an offset $\Delta x$. The refined bounding box parameter $x^{\prime}$ is then obtained by summing the offset $\Delta x$ with the initial value. Under this formulation, the decoder effectively performs a denoising task. This observation motivates us to explicitly introduce such a denoising task during training to facilitate learning. Specifically, we treat it as a shortcut to directly learn the relative offsets—bypassing the matching process altogether. In this denoising task, the decoder's input is generated by adding small perturbations around the ground-truth boxes to simulate noise, while the corresponding supervision signal is the ground-truth bounding box itself. This setup essentially provides prior knowledge of the anchor-to-ground-truth matching. The conclusions drawn in DN-DETR ~\cite{49} demonstrate that this strategy significantly enhances the robustness of the decoder in modeling bounding box coordinates, thereby improving overall detection accuracy.

\subsection{Loss Function} 
The loss function comprises three components: classification loss, bounding box regression loss, and denoising supervision loss. Its mathematical formulation is defined as follows:
\begin{equation}
L=\lambda_{c l s} L_{V F L}+\lambda_{b o x} L_{b o x}+\lambda_{d n} L_{d n}    
\end{equation}
To address the issues of class imbalance and the optimization of low-quality predicted boxes, we adopt Varifocal Loss (VFL) as the classification loss function. For the $i$-th predicted box, the classification loss is defined as:
\begin{equation}
\begin{aligned}
L_{V F L}= & -\sum_{i=1}^{N}\left\{y_{i} \cdot\left(1-p_{i}^{\gamma}\right) \log \left(p_{i}\right)\right. \\
& \left.+\left(1-y_{i}\right) \cdot p_{i}^{\gamma} \log \left(1-p_{i}\right)\right\}
\end{aligned}
\end{equation}
Among them, $y_{i}$ represents the true label (1 is a positive sample, 0 is background), ${p}_{i}$ is the category confidence of the predicted box, and $\gamma$ is the modulation factor (the default is $\gamma$ =2.0).

The bounding box regression loss is composed of a combination of the L1 loss and the Generalized Intersection over Union (GIoU) loss, which respectively constrain the absolute coordinate error and the geometric alignment accuracy:
\begin{equation}
\mathcal{G} L_{b o x}=\lambda_{L_{1}}\left\|b_{i}-\widehat{b}_{i}\right\|_{1}+\lambda_{G I O U}\left(1-\operatorname{GIOU}\left(b_{i}, \widehat{b}_{i}\right)\right)    
\end{equation}
Here, $b_{i}=\left(x_{i}, y_{i}, w_{i}, h_{i}\right)$ denotes the ground-truth box coordinates, and $\hat{b}_{i}$ represents the predicted box coordinates. By introducing the smallest enclosing region, the Generalized IoU (GIoU) effectively mitigates the gradient vanishing problem that arises in standard IoU when there is no overlap between predicted and ground-truth boxes. The weighting coefficients are determined via grid search and set as $\lambda_{\mathrm{L} 1}: \lambda_{\mathrm{GIoU}}=5: 2$, to balance absolute localization error and geometric consistency.

To enhance the robustness of the decoder against noise interference and to accelerate convergence, a denoising training mechanism is introduced. Specifically, zero-mean Gaussian noise $\epsilon \sim N\left(0, \sigma^{2} I\right)$ is added to the ground-truth box coordinates $b_{i}$ to generate a perturbed sample $\tilde{b}_{i}=b_{i}+\epsilon$, where the standard deviation $\sigma$ is set to 0.1 by default. The decoder is then explicitly trained to reconstruct the original coordinates from the noisy input:
\begin{equation}
L_{d n}=\sum_{i=1}^{M}\left\|D\left(\widetilde{b}_{i}\right)-b_{i}\right\|_{2}^{2}    
\end{equation}
Here, $\mathrm{D}(\cdot)$ denotes the coordinate mapping function of the decoder. By explicitly modeling localization uncertainty, this loss significantly improves the model’s localization stability in low-resolution and high-noise sonar imagery. The final weighting coefficients are set as $\lambda_{c l s}: \lambda_{b o x}: \lambda_{d n}=1: 2.5: 0.5$, and this configuration will be consistently adopted in the experimental evaluations presented in Section 4.

\section{Experiments}\label{sec4}
In this section, we first present the experimental setup and datasets used. Comprehensive benchmarking and comparative evaluations are then conducted on two large-scale datasets, URPC2021 ~\cite{50} and URPC2022 ~\cite{51}, including qualitative visualizations to demonstrate the superior performance of NAS-DETR. Subsequently, ablation studies are performed to investigate the impact of the search space configuration and the weighting of the differential entropy. Finally, to gain an intuitive understanding of how different parameters in the CNN-Transformer backbone affect differential entropy, we conduct a correlation analysis based on Spearman’s rank correlation coefficient. The details of the above components are provided in the following subsections.

\subsection{Dataset and implementation details} 
\subsubsection{Sonar Image Dataset} 
\begin{figure}
	\centering
\includegraphics[width=1.0\linewidth]{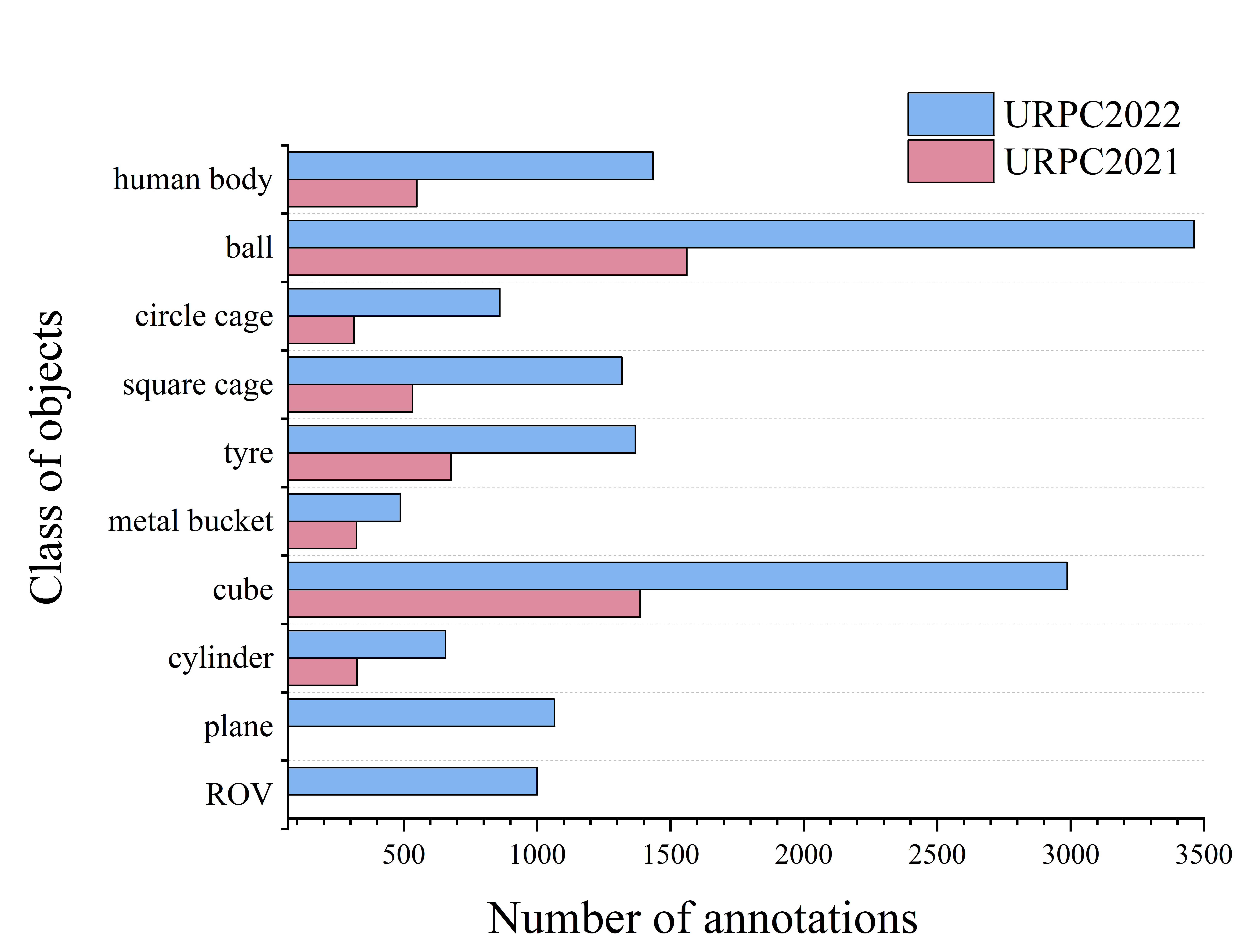}
	\caption{Distribution of different types of target samples on the URPC2021 and URPC2022 datasets.}\label{fig}
\end{figure}

In order to verify the performance of NAS-DETR in a high-noise complex environment closer to real deep-water conditions, we selected two representative opensource sonar object detection datasets in highly complex environments: URPC2021 and URPC2022.  
URPC2021 contains 6000 sonar images. The complete dataset has eight target categories, including human body, sphere, circular cage, squ-are cage, tire, metal barrel, cube and cylinder. In this work, according to the official division method and referring to the division principles of the standard VOC format dataset, we reconstructed the dataset with a training:testing ratio of 5000:1000. Each single image usually contains multiple instances and has an extremely uneven aspect ratio. We uniformly resize the images into rectangles to facilitate network training and inference.  
URPC2022 adds two categories, ROV and plane, to the base categories of URPC2021, with a total of 9200 images. Similarly to the division of URPC2021, the training set: test set is 8400:800, and the remaining settings are consistent with those of URPC2021. Figure.\ref{fig} shows the data distribution of the above two datasets. 

\subsubsection{Experiment settings} 
The experimental equipment and software used were as follows: Python 3.12 environment, PyTorch version 2.2.0. All architecture search processes were implemented on a computer equipped with a single NVIDIA GeForce RTX 3090 GPU, and the training phases of benchmark establishment, comparative experiments, and ablation experiments were all carried out on a server equipped with two NVIDIA GeForce RTX 3090 GPUs. The establishment of the benchmark was implemented based on the MMdetection framework, and the training parameters were all based on the default settings of the framework, including data augmentation operations such as random photometric distortion and random horizontal flip. During the training process of NAS-DETR, the learning rate was constantly set to 0.0001, and data augmentation methods such as RandomPhotometricDistort, RandomZoomOut, and RandomIoUCrop were used.  
\subsubsection{NAS} 
For the NAS process, the Block types are fixed throughout the search phase, as given in Table \ref{tab:t2-t3}. In Table \ref{tab:t2-t3}, the settings for \( A_1 = \{0, 0, 1, 1, 2, 4\} \) and \( A_2 = \{0, 0, 1, 1, 3, 6\} \) are presented with the corresponding data separated by slashes in each cell for easy comparison. We adopt different mutation strategies for Blocks of different Stages, as shown in Table\ref{tab:4}. The entire search process proceeds for 20,000 rounds, and at the end of evolution, the backbone with the highest fitness function value is returned. The NAS process is carried out under two entropy - weight settings, namely \( A_1 = \{0, 0, 1, 1, 2, 4\} \) and \( A_2 = \{0, 0, 1, 1, 3, 6\} \). We conduct experiments on these two groups of settings separately in all subsequent experiments. The final architectures obtained under the two settings are shown in Table \ref{tab:t2-t3}. To evaluate the detection performance of all methods, we employed multiple object detection metrics, including mmAP, mAP50, mAP75, and mAP50 for each category.

\begin{figure*}
	\centering
	\includegraphics[width=0.95\linewidth]{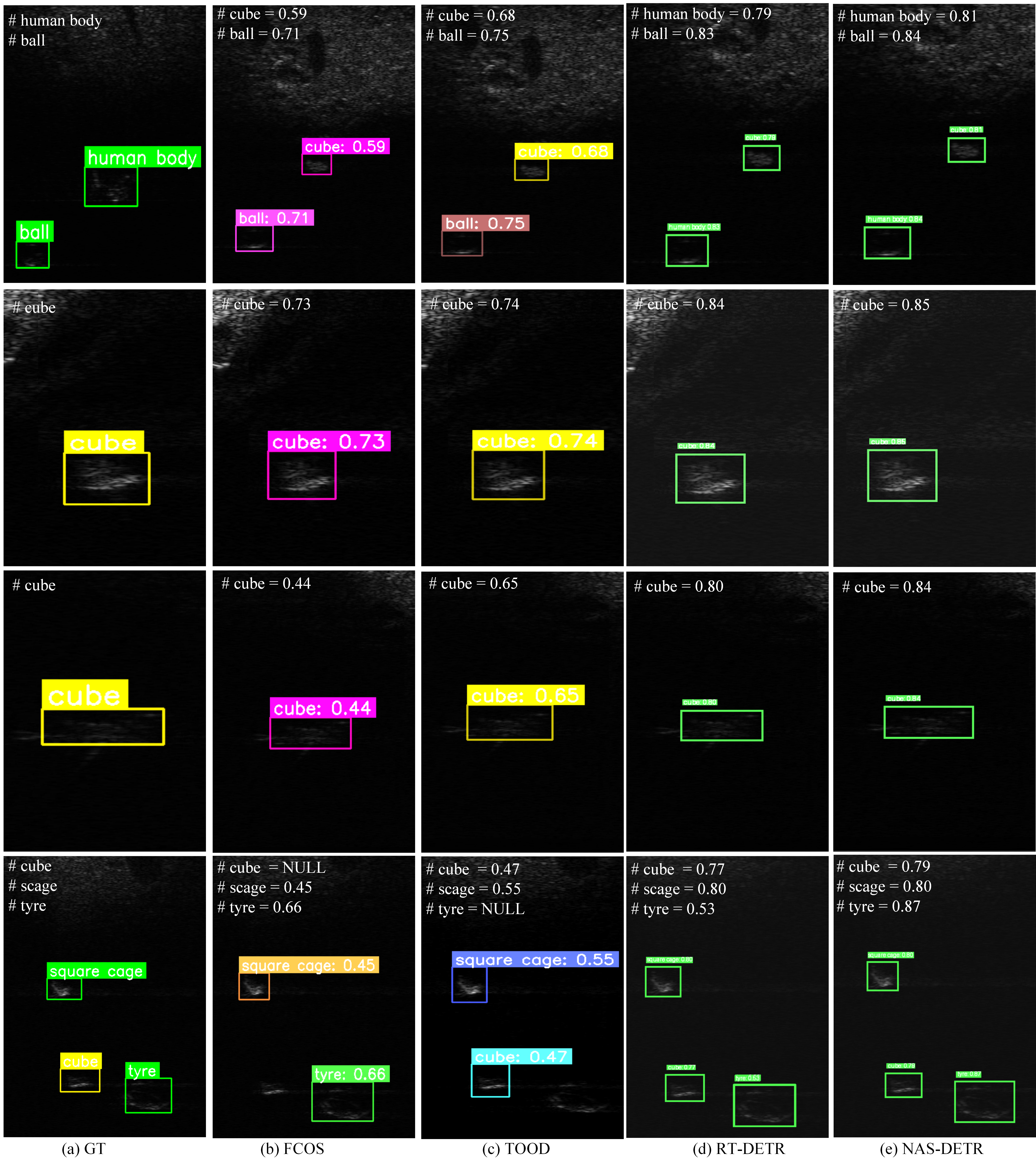}
	\caption{Visualization of detection results of NAS-DETR and methods in the benchmark on URPC2021.}\label{fig7}
\end{figure*}

\begin{table*}
\caption{Architecture search results under differential entropy weight settings ($A_1$/$A_2$ values in each cell,the left data represents the results under the A1 parameter setting, and the right data corresponds to the A2 parameter setting. * denotes ResBlock with max-pooling).}  
\label{tab:t2-t3}
\centering
\resizebox{\textwidth}{!}{%
\begin{tabular}{lccccccccc}
\toprule
Block                                & kernal & in   & out  & stride & bottleneck & layers & Hidden\_dim & Dim\_feedforward & level \\ 
\midrule
ResBlock*                            & 3/3    & 3/3  & 32/32 & 4/4    & 32/32      & 1/1    & -/-         & -/-             & C1    \\
ResBlock                             & 5/5    & 32/32& 128/112 & 1/1    & 40/48      & 3/3    & -/-         & -/-             & C2    \\
ResBlock                             & 5/5    & 128/112 & 448/448 & 2/2    & 80/72      & 8/8    & -/-         & -/-             & C3    \\
ResBlock                             & 5/5    & 448/448 & 1280/1024 & 2/2    & 128/104    & 10/10  & -/-         & -/-             & C4    \\
ResBlock                             & 5/5    & 1280/1024 & 2048/2000 & 2/2    & 240/304    & 10/10  & -/-         & -/-             & C5    \\
\multicolumn{1}{l}{TransformerBlock} & -/-     & 2048/2000 & 256/256 & -/-     & -/-        & 1/1    & 424/504     & 912/1024        & C6    \\ 
\bottomrule                                
\end{tabular}
}
\end{table*}

\begin{table*}[h]
\centering
\caption{Mutation operations for each module, where * has the same meaning as described above }
\resizebox{0.9\textwidth}{!}{\begin{tabular}{lcccccc}
\toprule
Mutation & Kernel & Layer & Channel & bottleneck & Hidden & Feedforward \\
  type &  &  &  & width & Dimension & Dimension \\
\midrule
Mutation & Choose & $\pm$ & $\times$ & $\times$ & $\pm$ & $\pm$ \\
operation & $[3,5]$ & $[1,2]$ & $[1.5,1.25,0.8,0.6,0.5]$ & $[1.5,1.25,0.8,0.6,0.5]$ & $[8,16,32,64,128]$ & $[8,16,32]$ \\
Block & ResBlock & ResBlock & ResBlock, ResBlock* & ResBlock, ResBlock* & Transformer Block & Transformer Block \\
\bottomrule
\end{tabular}}
\label{tab:4}
\end{table*}

\begin{figure*}
	\centering
	\includegraphics[width=0.95\linewidth]{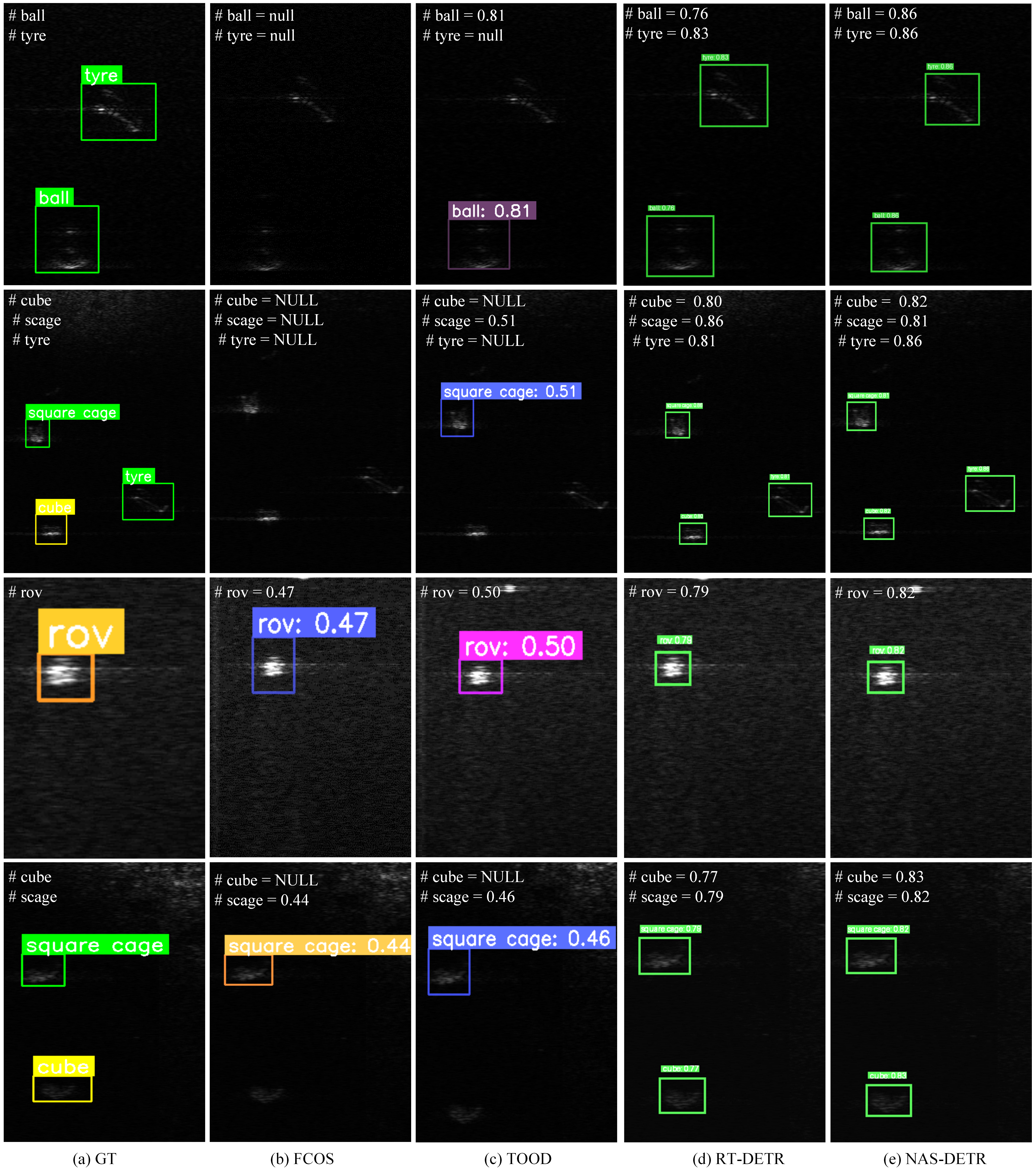}
	\caption{Visualization of detection results of NAS-DETR and methods in the benchmark on URPC2022.}\label{fig8}
\end{figure*}

\subsection{Results on URPC2021 Dataset} 
Table \ref{tab:t4} and Table \ref{tab:t5} show the experimental results of various object detection methods on the URPC2021 dataset. Experimental results show that the NAS-DETR (A1 parameter setting) model achieves state-of-the-art(SOTA) results under various indicators. Its mmAP reaches 0.538, not only surpassing 0.525 of RT-DETR, but also significantly leading models such as DDOD (0.4519) and VFNet (0.4501). The mAP50 index is 0.919, much higher than 0.744 of TOOD and 0.806 of FCOS (101), and in mAP75 it is 0.569, an increase of 0.032 compared with 0.537 of RT-DETR, and greatly leading RTMDet (0.473) and VFNet (0.445). Under strict positioning criteria, its precise positioning ability has significant advantages, and it can more accurately determine the target position. Moreover, in the mAP50 detection of various categories, although individual categories have different levels, most categories are leading as a whole. The results show that the NAS-DETR model has outstanding advantages, fully reflecting its detection generalization ability for different categories of targets.

Figure~\ref{fig7} shows the visual results of object detection in the URPC2021 dataset. Among these algorithms, NAS-DE-TR(A2) demonstrates the most excellent detection effect, followed by RT-DETR, while TOOD and FCOS perform relatively poorly. Specifically, both TOOD and FCOS have missed detection in the fourth group of detection results, indicating that they have certain limitations in detecting targets and cannot fully identify all targets. Moreover, in the first group of results, these two algorithms also misidentify the human body as a cube. This misjudgment fully reflects their insufficient ability in feature distinction and category recognition, and they cannot accurately extract and classify the features of different targets. By contrast, both NAS-DETR(A2) and RT-DETR can identify targets more accurately during the detection process. Further comparison shows that the confidence level of NAS-DETR(A2) is generally higher than that of RT-DETR. A higher confidence level means that the detection results of NAS-DETR(A2) are more reliable. In practical application scenarios, it is more confident to ensure the accuracy of detection results, effectively reducing the risk of misjudgment and missed detection, thus further highlighting the significant advantages of NAS-DETR(A2) in object detection tasks. 

\begin{table}
\setlength\tabcolsep{4mm}
\caption{Overall Detection Performance Metrics of URPC2021 Dataset}
\label{tab:t4}
\resizebox{0.45\textwidth}{!}{
\begin{tabular}{lccc}
\toprule
\multirow{2}{*}{Methods} & \multicolumn{3}{c}{metric} \\
\cmidrule{2 - 4}
       & mmap  & map50 & map75 \\
\midrule
TOOD      & 0.3915& 0.744 & 0.342 \\
FCOS(101) & 0.4219& 0.806 & 0.370  \\
RTMDet    & 0.4772& 0.851 & 0.473 \\
DDOD      & 0.4519& 0.816 & 0.422 \\
RT-DETR   & 0.5250& 0.911 & 0.537 \\
VFNet     & 0.4501& 0.804 & 0.445 \\
NAS-DETR(A1) & \textbf{0.5380}& \textbf{0.919} & \textbf{0.569} \\
\bottomrule
\end{tabular}
}
\end{table}

\begin{table*}
\caption{Detection Performance Metrics for Each Category (map50) of URPC2021 Dataset}
\label{tab:t5}
\resizebox{0.9\textwidth}{!}{\begin{tabular}{lcccccccc}
\toprule
\multirow{2}{*}{Methods} & \multicolumn{8}{c}{map50} \\
\cmidrule{2 - 9} & ball & square cage & tyre & metal bucket & human body & cube & circle cage & cylinder \\
\midrule
TOOD & 0.891 & 0.805 & 0.665 & 0.714 & 0.757 & 0.864 & 0.704 & 0.554 \\
FCOS(101) & 0.897 & 0.822 & 0.834 & 0.879 & 0.8 & 0.869 & 0.855 & 0.49 \\
RTMDet & 0.898 & 0.842 & 0.883 & 0.867 & 0.889 & 0.89 & 0.85 & 0.689 \\
DDOD & 0.889 & 0.797 & 0.819 & 0.879 & 0.883 & 0.886 & 0.872 & 0.507 \\
RT-DETR & 0.966 & 0.905 & 0.887 & \textbf{0.905} & \textbf{0.959} & 0.95 & 0.899 & \textbf{0.82} \\
VFNet & 0.889 & 0.797 & 0.78 & 0.853 & 0.883 & 0.893 & 0.877 & 0.46 \\
NAS-DETR(A1) & \textbf{0.967} & \textbf{0.927} & \textbf{0.913} & 0.897 & 0.953 & \textbf{0.971} & \textbf{0.906} & 0.816 \\
\bottomrule
\end{tabular}}
\end{table*}

\subsection{Results on URPC2022 Dataset} 

Table \ref{tab:t6} and Table \ref{tab:t7} present the experimental results of various object detection methods on the URPC2022 dataset. In the sonar object detection task of the URPC2022 dataset, the NAS-DETR (A2) model demonstrates significant advantages, with extremely outstanding overall detection performance. Its mmAP reaches 0.492, improving by 0.016 and 0.0278 compared to RT-DETR (0.476) and DDOD (0.4642), respectively. The mAP50 is 0.954, and the mAP75 is 0.447, showing excellent performance under different strict localization requirements and superior target detection capabilities. Additionally, in the mAP50 statistics for different object categories, NAS-DETR (A2) achieves 0.948 in the ball category, outperforming RT-DETR's 0.941. It also exhibits excellent performance in categories such as plane,circle cage, and rov (e.g., reaching 1 for plane and 0.951 for "circle cage"). Although it slightly lags in some categories, NAS-DETR maintains an overall leading position and exhibits outstanding generalization ability.

Figure~\ref{fig8} shows the the visualization results of object detection on the URPC2022 dataset, further validate the quantitative conclusions. FCOS and TOOD reveal significant missed - detection issues. For example, in the detection of sonar images in the first and second rows, FCOS misses all the targets. Meanwhile, the confidence levels of targets for both FCOS and TOOD are generally at a low level, which means the reliability of their detection results is poor. NAS-DETR and RT-DETR demonstrate more excellent detection effects. Especially NAS-DETR, its detection confidence is significantly higher than other methods, intuitively reflecting that its detection results have better reliability and stability in practical applications, effectively reducing the risk of missed judgments and highlighting its advantages in object - detection tasks. 

\begin{table}
\setlength\tabcolsep{4mm}
\caption{Overall Detection Performance Metrics of URPC2022 Dataset}
\label{tab:t6}
\resizebox{0.45\textwidth}{!}{\begin{tabular}{lccc}
\toprule
\multirow{2}{*}{Methods} & \multicolumn{3}{c}{metric} \\
\cmidrule{2 - 4}
       & mmap  & map50 & map75 \\
\midrule
TOOD      & 0.3665 & 0.801 & 0.262 \\
FCOS(101) & 0.4523 & 0.904 & 0.357 \\
NAS-FCOS  & 0.3325 & 0.806 & 0.200   \\
RTMDet    & 0.4355 & 0.88  & 0.351 \\
DDOD      & 0.4642 & 0.907 & 0.420  \\
RT-DETR   & 0.476  & 0.951 & 0.419 \\
NAS-DETR(A2) & \textbf{0.492} & \textbf{0.954} & \textbf{0.447} \\
\bottomrule
\end{tabular}}

\end{table}

\begin{table*}
\caption{Detection Performance Metrics for Each Category (map50) of URPC2022 Dataset}
\label{tab:t7}
\resizebox{0.9\textwidth}{!}{\begin{tabular}{lccccccccc}
\toprule
\multirow{2}{*}{Methods}  & \multicolumn{9}{c}{map50} \\
\cmidrule{2 - 10}
       & ball & square cage & tyre & metal bucket & human body & cube & plane & circle cage & rov \\
\midrule
TOOD & 0.84 & 0.751 & 0.812 & 0.833 & 0.844 & 0.742 & 0.908 & 0.609 & 0.869 \\
FCOS(101) & 0.883 & 0.893 & 0.903 & \textbf{1} & 0.892 & 0.856 & 0.909 & 0.931 & 0.908 \\
NAS-FCOS & 0.857 & 0.832 & 0.865 & \textbf{1} & 0.872 & 0.749 & 0.831 & 0.699 & 0.857 \\
RTMDet & 0.878 & 0.906 & 0.905 & 0.883 & 0.876 & 0.885 & 0.986 & 0.853 & 0.875 \\
DDOD & 0.877 & 0.857 & 0.907 & \textbf{1} & 0.905 & 0.884 & 0.909 & 0.875 & 0.906 \\
RT-DETR & 0.941 & 0.89 & 0.956 & \textbf{1} & \textbf{0.937} & \textbf{0.893} & 0.999 & 0.927 & 0.988 \\
NAS-DETR(A2) & \textbf{0.948} & \textbf{0.922} & \textbf{0.963} & \textbf{1} & 0.909 & 0.858 & \textbf{1} & \textbf{0.951} & \textbf{1} \\
\bottomrule
\end{tabular}}
\end{table*}

\subsection{Ablation experiment}
We compare the performance of RT-DETR, MAE-DETR, and NAS-DETR with two parameter settings on the URPC series datasets to verify the applicability of the proposed NAS method as our ablation experiment. Among them, MAE-DETR is obtained by replacing the Backbone architecture of RT-DETR with the Backbone architecture searched in MAE-DET. The experimental results are shown in Table \ref{tab:t8}  and Table \ref{tab:t9}. It can be seen that the initial RT-DETR has the worst performance. After replacing its Backbone with MAE Backbone, there are \(0.1\%\) and \(0.2\%\) improvements in mmap on URPC2021 and URPC2022 respectively, which fully shows that the Zero-shot NAS method based on the maximum entropy principle is not only effective for traditional optical images but also applicable to sonar data. After introducing our NAS-DETR method, there are \(1.2\%\) and \(1.4\%\) performance improvements respectively compared with MAE-DETR. The magnitude of which fully demonstrates the performance - gain effect of introducing the search operations for Transformer Block (including numerical scale scaling, differential entropy estimation, etc.) on the final architecture.

\begin{table*}
\caption{Ablation experiment results on the URPC2021 dataset.}
\label{tab:t8}
\resizebox{0.9\textwidth}{!}{
\setlength\tabcolsep{5mm}
\begin{tabular}{lcccccc}
\toprule
\multirow{2}{*}{Methods} & \multicolumn{3}{c}{metric} & & \multicolumn{2}{c}{map50} \\
\cmidrule{2 - 4} \cmidrule{6-7}
       & mmap  & map50 & map75  &  & ball & Square cage \\
\midrule
RT-DETR  & 0.525 & 0.911 & 0.537 & & 0.966 & 0.905     \\
MAE-DETR & 0.526 & 0.917 & 0.546 & & \textbf{0.97} & 0.915   \\
NAS-DETR(A2) & 0.526 & 0.915 & 0.555 & & \textbf{0.97} & 0.921       \\
NAS-DETR(A1) & \textbf{0.538} & \textbf{0.919} & \textbf{0.569} & & 0.967 & \textbf{0.927}       \\
\bottomrule
\end{tabular}
}

\resizebox{0.9\textwidth}{!}{
\setlength\tabcolsep{4mm}
\begin{tabular}{lcccccc}
\toprule
\multirow{2}{*}{Methods} & \multicolumn{6}{c}{map50} \\
\cmidrule{2 - 7}
& tyre & metal bucket & human body & cube & circle cage & cylinder \\
\midrule
RT-DETR  & 0.887 & 0.905 & \textbf{0.959} & 0.95 & 0.899 & 0.82    \\
MAE-DETR & 0.904 & 0.898 & 0.958 & 0.957 & 0.886 & \textbf{0.848}  \\
NAS-DETR(A2) & \textbf{0.928} & \textbf{0.908} & 0.955 & 0.956 & 0.888 & 0.791      \\
NAS-DETR(A1) & 0.913 & 0.897 & 0.953 & \textbf{0.971} & \textbf{0.906} & 0.816       \\
\bottomrule
\end{tabular}
}

\end{table*}

\begin{table*}
\caption{Ablation experiment results on the URPC2022 dataset.}
\label{tab:t9}
\resizebox{0.9\textwidth}{!}{
\setlength\tabcolsep{5mm}
\begin{tabular}{lccccccc}
\toprule
\multirow{2}{*}{Methods} & \multicolumn{3}{c}{metric} & & \multicolumn{2}{c}{map50} \\
\cmidrule{2 - 4} \cmidrule{6-8}
       & mmap  & map50 & map75  &  & ball & square cage & tyre \\
\midrule
RT-DETR  & 0.476 & 0.951 & 0.419 & & 0.941 & 0.89 & 0.956     \\
MAE-DETR & 0.478 & 0.951 & 0.398 & & \textbf{0.949} & 0.89 & 0.959  \\
NAS-DETR(A2) & 0.488 & \textbf{0.957} & 0.425 & & \textbf{0.949} & 0.905 & \textbf{0.969}       \\
NAS-DETR(A1) & \textbf{0.492} & 0.954 & \textbf{0.447} & & 0.948 & \textbf{0.922} & 0.963      \\
\bottomrule
\end{tabular}
}
\resizebox{0.9\textwidth}{!}{
\setlength\tabcolsep{3.5mm}
\begin{tabular}{lccccccc}
\toprule
\multirow{2}{*}{Methods} & \multicolumn{7}{c}{map50} \\
\cmidrule{2 - 8}
& metal bucket & human body & cube & plane & circle cage & cylinder & rov \\
\midrule
RT-DETR  & \textbf{1} & 0.937 & \textbf{0.893} & 0.999 & 0.927 & 0.978 & 0.988 \\
MAE-DET  & \textbf{1} & 0.938 & 0.883 & \textbf{1} & 0.935 & 0.969 & 0.99 \\
NAS-DETR(A1)  & \textbf{1} & \textbf{0.948} & 0.875 & \textbf{1} & \textbf{0.959} & 0.965 & 0.998 \\
NAS-DETR(A2)  & \textbf{1} & 0.909 & 0.858 & \textbf{1} & 0.951 & \textbf{0.994} & \textbf{1}      \\
\bottomrule
\end{tabular}
}

\end{table*}

In summary, a guiding conclusion can be drawn: the architecture optimized by the NAS method based on the principle of maximum entropy for differential entropy outperforms the initial architecture, and the final performance is directly proportional to the architecture scope searched by NAS. 

\subsection{Computational efficiency analysis}
Additionally, we also compared the FPS and GFLOPS of model detection, as shown in Table \ref{tab:t10}. It can be seen from the table that NAS-DETR achieves a good balance in detection accuracy, speed, and computational cost. The FPS of NAS-DETR(A1) and NAS-DETR(A2) are 73.8 and 71.2 respectively. Although lower than that of RT-DETR (114.4FPS), compared with other methods, the FPS has a huge leading margin. Their GFLOPS are approximately 145. NAS-DETR can ensure a significant advantage in detection accuracy without excessively increasing the computational cost, avoiding the problem of excessive consumption of computational resources due to the pursuit of high accuracy, and the comprehensive performance is outstanding.  

\begin{table}
\setlength\tabcolsep{5mm}
\caption{Computational efficiency of different methods}
\label{tab:t10}
\resizebox{0.45\textwidth}{!}{\begin{tabular}{lcc}
\toprule
method & FPS(frame/s) & GFLOPS(G) \\
\midrule
TOOD & 39.37 & 144.364 \\
FCOS(101) & 34.25 & 218.005 \\
NAS-FCOS & 65.36 & \textbf{103.793}\\
RTMDet & 17.24 & 122.696 \\
DDOD & 27.39 & 142.361 \\
RT-DETR & \textbf{114.4} & 149.771 \\
NAS-DETR(A1) & 73.8 & 145.815 \\
NAS-DETR(A2) & 71.2 & 144.758 \\
\bottomrule
\end{tabular}}

\end{table}

After quantizing NAS-DETR using Tensor-RT, the model inference efficiency is greatly improved. The FPS of NAS-DETR(A1) and NAS-DETR(A2) reach 294.9 and 288 respectively, achieving a nearly three-fold performance improvement compared with before quantization. This result not only fully verifies the excellent potential of NAS-DETR in inference acceleration but also demonstrates the great potential of NAS-DETR for deployment in industrial scenarios. 

\subsection{Correlation analysis}\label{sec:4.6}
In a bid to introduce this NAS method more credibly, we adopt the Spearman correlation coefficient, which is a credible way to analyze the relationship between the parameters and the differential entropy in CNN-Transformer Backbone in this section.

\begin{figure*}
	\centering
	\includegraphics[width=0.95\linewidth]{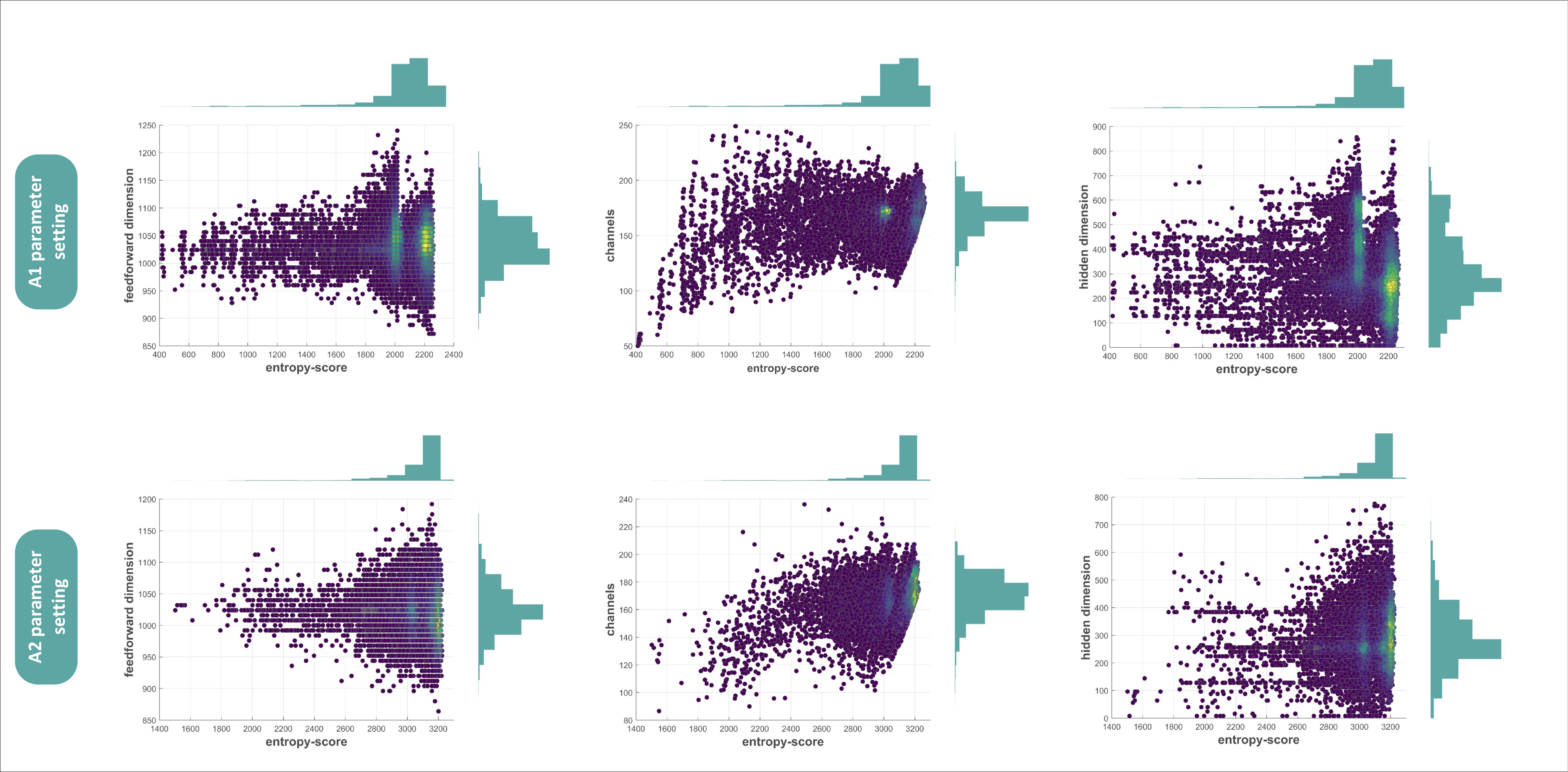}
	\caption{Sampling distribution of architecture parameter - differential entropy data pairs for various network architectures.}\label{fig:9}
\end{figure*}

With our best effort to minimize the effects of randomness in our results, we collected a substantial dataset of paired structure parameters - differential entropy values. Notably, the differential entropy was calculated using Eq\ref{eq:29} to ensure that the analysis directly reflects the algorithmic behavior of the search process. We designed two experimental groups corresponding to the two sets of weights (A1 and A2) proposed in Section 4.5. Under each experimental setting, we collected one hundred thousand data pairs, as illustrated in Figure.\ref{fig:9}.

\subsubsection{Definition of the average architecture control parameter.} 
Direct analysis of the correlation between all parameters and differential  entropy would lead to computational complexity and confusing results. To address this, we propose a simplified approach based on parameter averaging. However, traditional arithmetic averaging fails to accommodate the "extensive property" characteristic of differential entropy (which increases linearly with network depth L, as shown in Eq\ref{eq:22}). Inspired by thermodynamics, if we consider the mean value as an "intensive property," then its product with network size should characterize an "extensive property." Consequently, we redefine the calculation rules for average parameters to ensure strict correspondence with the extensivity of differential entropy.

Consider an L-layer CNN with its each layer has different parameters. For the complete architecture, each parameter corresponds to an average value, such as channel count and kernel size. If we then create another CNN structure where all layers have parameters equal to these average values, according to Eq\ref{eq:14}:

\begin{equation}
\left\{\begin{aligned}
\boldsymbol{H}_{\boldsymbol{G}_{a}} & =\sum_{i=1}^{\boldsymbol{L}} \log \left(c_{i} k_{i}^{2}\right) \\
\boldsymbol{H}_{\boldsymbol{G}_{b}} & =\boldsymbol{L} \boldsymbol{l o g}\left(\overline{\boldsymbol{k}}^{2} \boldsymbol{c}_{\boldsymbol{i}}\right)
\end{aligned}\right.
\end{equation}

According to the property of the mean value:

\begin{equation}
\boldsymbol{H}_{\boldsymbol{G}_{a}}=\boldsymbol{H}_{\boldsymbol{G}_{b}}
\end{equation}

Therefore, for the definition formula of the mean of $\bar{c}$ and $\bar{k}$, it is not difficult to find that this is a geometric mean:  

\begin{equation}
\left\{\begin{array}{l}
\bar{k}=\exp \left(\frac{1}{L} \sum_{i=1}^{L} \log \left(\bar{k}_{i}\right)\right) \\
\bar{c}=\exp \left(\frac{1}{L} \sum_{i=1}^{L} \log \left(\bar{c}_{i}\right)\right)
\end{array}\right.
\end{equation}

\subsubsection{Spearman's rank correlation coefficient} 
To investigate the correlation between architectural parameters of CNN - Transformer backbone networks and differential entropy, this study employs Spearman's Rank Correlation Coefficient \cite{52} for non-parametric statistical testing. 
For the parameter set \(X = \{x_1,x_2,\ldots,x_n\}\) and differential entropy set \(Y=\{y_1,y_2,\ldots,y_n\}\), we have:

\begin{equation}
\rho=1-\frac{6 \sum_{\mathrm{i}=1}^{\mathrm{n}} \mathrm{~d}_{\mathrm{i}}^{2}}{\mathrm{n}\left(\mathrm{n}^{2}-1\right)}
\end{equation}

 \(d_i = R(x_i)-R(y_i)\), where \(n\) is the sample size. The Spearman's rank correlation coefficient \(\rho\) satisfies \(\rho\in[- 1,1]\). 
\begin{itemize}
    \item  \(\rho>0\), there is a positive monotonic correlation.
    \item  \(\rho < 0\), there is a negative monotonic correlation.
\end{itemize}

\subsubsection{Correlation Analysis}\label{4.6.3}
We conduct a Spearman correlation analysis on the data presented in Figure.\ref{fig:9}.Table\ref{tab:t11} demonstrates significant patterns under two parameter settings (A1/A2): First, network depth \(L\) exhibits a strong positive correlation with differential entropy (A1: \(\rho = 0.904\), A2: \(\rho = 0.826\)), confirming the critical role of depth in feature representation. Second, average channel count (A1: \(\rho = 0.247\rightarrow\)A2: \(\rho = 0.443\)) and convolution kernel size (A1: \(\rho = 0.291\rightarrow\)A2: \(\rho = 0.437\)) both show significant positive correlations, with enhanced effects under high - entropy weights, indicating that width parameters increase entropy values through cooperative mechanisms.
However, Transformer parameters present contradictory results: under A1 settings, \(\text{hidden\_dim}\) shows no significant association with \(\text{Score}\) (\(\rho=-0.01\)), while \(\text{feedforward}\) demonstrates significant negative correlation (\(\rho = 0.904\)); under A2 settings, \(\text{hidden\_dim}\) shifts to a weak negative correlation (\(\rho=-0.165\)), and \(\text{feedforward}\) exhibits a low positive correlation (\(\rho = 0.205\)). This anomaly stems from computational resource competition between Transformer and CNN modules under Flops constraints rather than contradicting the conclusions in Eq.\ref{eq:14}. Through control experiments with fixed CNN architecture, \(\text{hidden\_dim}\) (\(\rho = 0.693\)) and \(\text{feedforward}\) (\(\rho = 0.647\)) display significant positive correlations with \(\text{Score}\), confirming their global feature modeling capabilities.



\begin{table*}[htb]
    \caption{Correlation analysis under A1 differential entropy weight settings.(\cellcolor[HTML]{F7F7F7}{\color[HTML]{111111} Note: ***, **, and * represent the significance levels of 1\%, 5\%, and 10\%, respectively})}
    \centering
    \label{tab:t11}
    \begin{tabular*}{0.95\textwidth}{@{\extracolsep{\fill}}lccccc}
    \toprule
    Parameters & L & Transformer\_btn & avg\_channels & avg\_k & hdn \\ \midrule
    Score(A1) & 0.904*** & -0.01*** & 0.247*** & 0.291*** & -0.287*** \\
    Score(A2) & 0.826*** & -0.165*** & 0.443*** & 0.437*** & 0.205*** \\ \bottomrule
    \end{tabular*}
\end{table*}

\subsubsection{Correlation Analysis Under Bivariate Conditions} 
In the previous section, our correlation analysis experiments were based on data sampled from the search process, which imposed maximum constraints on Flops. While this approach is well suited for CNN layers, it introduces a challenge for Transformer Blocks: the previous correlation analysis was limited by the sampling strategy under total Flops constraints: the high computational demands of individual Transformer modules competed with CNN computational resources, potentially increasing local differential entropy while reducing global entropy values.

Therefore, to eliminate this bias, our study fixed the CNN architecture (as specified in Table\ref{tab:t2-t3}, Table\ref{tab:t3}), systematically traversed the Transformer parameters, and calculated the entropy values. Through uniform sampling, we eliminated mutual information interference between parameters, more accurately reflecting global structural correlations.

Similarly to Section \ref{4.6.3} we collected experimental samples for correlation analysis under two differential entropy weighting methods \(A_1, A_2\). For parameter \(d_{\text{model}}\), we provide the sample set \(S_{\text{model}}=\{X|512\geq X\geq256, X\bmod 8 = 0\}\). Similarly, for the parameter \(d_{\text{feedforward}}\), we have \(S_{\text{feedforward}}=\{X|2048\geq X\geq512, X\bmod 12 = 0\}\). Each sampling selects one combination of \(d_{\text{model}}, d_{\text{feedforward}}\) from \(S_{\text{model}}, S_{\text{feedforward}}\), respectively. If we define the mapping relationship between total architectural differential entropy \(E_{\text{total}}\) and \(d_{\text{model}}, d_{\text{feedforward}}\) as \(E_{\text{total}} = h(d_{\text{model}})\) and \(E_{\text{total}}=h(d_{\text{feedforward}})\), then the sample set of parameter differential entropy parameter pairs can be given as Eq\ref{eq39}:
\begin{equation}
\begin{aligned}
D = \{&(d_{\text{model}},d_{\text{feedforward}},E_{\text{total}})\\
&|d_{\text{model}}\in S_{\text{model}},d_{\text{feedforward}}\in S_{\text{feedforward}}\}
\end{aligned}
\label{eq39}
\end{equation}

The analysis results are shown in Table\ref{tab:t12}.


\begin{table}[htb]
    \centering
    \caption{Correlation analysis under A2 differential entropy weight settings. (\cellcolor[HTML]{F7F7F7}{\color[HTML]{111111}Note: ***, **, and * represent the significance levels of 1\%, 5\%, and 10\%, respectively})}
    \label{tab:t12}
    \begin{tabular*}{0.9\linewidth}{@{\extracolsep{\fill}}lcc}
        \toprule
        Parameters & hdn      & btn      \\ \midrule
        Score(A1)  & 0.693*** & 0.655*** \\
        Score(A2)  & 0.647*** & 0.697*** \\ \bottomrule
    \end{tabular*}
\end{table}

\section{Conclusion}\label{sec5}
The low resolution and high noise characteristics of sonar images, coupled with the limited local receptive field of traditional CNN models, severely constrain underwater target detection performance. Addressing this challenge, this paper proposes an end-to-end detection framework, NAS-DETR, which is optimized through Neural Architecture Search(NAS), achieving efficient and robust detection for sonar images. Through systematic experiments and theoretical analysis, the main contributions and conclusions of this study are as follows:  

First, based on the maximum entropy principle for mixed backbone search, we propose an improved Zero-shot NAS method that maximizes neural network differential entropy to identify an efficient CNN-Transformer hybrid backbone network. This architecture combines CNN's local feature extraction capabilities with Transformer's global contextual modeling advantages, significantly enhancing feature representation for sparse targets in sonar images. Subsequently, we develop a robust decoder design. Through a content-position decoupled Query initialization strategy and a hybrid loss function with multi-task collaborative optimization (integrating classification loss, bounding box regression loss, and denoising supervision loss), we effectively mitigate traditional DETR's slow convergence and insufficient localization precision issues, while demonstrating enhanced stability in high-noise environments.  

Experiments on URPC2021 and URPC2022 datasets show that NAS-DETR achieves mmAP of 0.538 and 0.492, respectively—improvements of 2.5\% and 3.4\% over RT-DETR, while maintaining 71.2–73.8 FPS, balancing detection accuracy and real-time performance. Further Spearman correlation analysis reveals positive correlations between backbone network depth, channel count, convolution kernel size, and differential entropy (\(\rho = 0.826\text{–}0.904\)), providing interpretable theoretical support for the NAS method. Notably, after Tensor-RT quantization, NAS-DETR's inference speed increases to 288–294.9 FPS, validating its potential for efficient deployment in industrial scenarios.  

Our future work will focus on lightweight search strategies, multi-modal data fusion, and adaptive detection in dynamic environments to further enhance model generalization and practicality in complex underwater tasks.  

\section*{Acknowledgments}
This work was supported by the National Natural Science Foundation of China under Grant 52401362. 
{
\fontsize{8.2pt}{9.84pt}\selectfont

\bibliographystyle{unsrt}
\bibliography{regbib}
}

\end{document}